\documentclass{article}

\usepackage{microtype}
\usepackage{graphicx}
\usepackage{grffile} % needed for the graphicx pkg to check for known extensions (option multidot, enabled by default)

\usepackage{booktabs} % for professional tables

\usepackage{hyperref}

\newcommand{\icml}{}

 \usepackage[accepted]{icml2023}

\usepackage{amsmath}
\usepackage{amssymb}
\usepackage{mathtools}
\usepackage{amsthm}

\usepackage[capitalize,noabbrev]{cleveref}

\theoremstyle{plain}
\newtheorem{theorem}{Theorem}[section]

\newtheorem{lemma}[theorem]{Lemma}
\newtheorem{corollary}[theorem]{Corollary}
\theoremstyle{definition}
\newtheorem{definition}[theorem]{Definition}

\theoremstyle{remark}

\usepackage[textsize=tiny]{todonotes}

\usepackage{amsmath,amsfonts,bm}

\def\eqref#1{equation~\ref{#1}}
\def\1{\bm{1}}

\newcommand{\test}{\mathcal{D_{\mathrm{test}}}}

\def\vs{{\bm{s}}}

\DeclareMathAlphabet{\mathsfit}{\encodingdefault}{\sfdefault}{m}{sl}
\SetMathAlphabet{\mathsfit}{bold}{\encodingdefault}{\sfdefault}{bx}{n}

\DeclareMathOperator*{\argmax}{arg\,max}

\usepackage{url}

\usepackage[utf8]{inputenc} % allow utf-8 input
\usepackage[T1]{fontenc}    % use 8-bit T1 fonts

\usepackage{amsfonts}       % blackboard math symbols
\usepackage{nicefrac}       % compact symbols for 1/2, \etc
\usepackage{xcolor}         % colors
\usepackage{comment}
\usepackage{sidecap}
\usepackage{subcaption}

\usepackage[toc,page,header]{appendix}
\usepackage{minitoc}

\usepackage{natbib}

\usepackage{epsfig}
\usepackage{graphicx}
\usepackage{algorithmic}

\usepackage[algo2e,lined,boxed,commentsnumbered,ruled]{algorithm2e}

\usepackage{multirow}
\usepackage{multicol}
\usepackage{amssymb,amsmath,amsopn}
\usepackage{shadethm}
\usepackage{booktabs}
\usepackage{color}
\usepackage{url}
\usepackage{float}
\usepackage{threeparttable}
\usepackage{array}
\usepackage{cases}
\usepackage{colortbl}
\usepackage[tableposition=top, labelfont={it}, labelsep=period]{caption}
\usepackage{array}
\usepackage{enumitem}
\usepackage{footnote}

\usepackage{xcolor}

\ifdefined\icml
\else
    \usepackage{floatrow}
\fi

\usepackage{mathrsfs}

\usepackage{amssymb,amsmath,amsopn,amsthm}

\usepackage{commath}
\usepackage{mathtools}

\newcommand{\Lnorm}[1]{\ell^{#1}}

\def \real{\mathbb R}
\def \natural{\mathbb N}

\def \funcStep{W}

\newfont{\bboard}{msbm10 scaled\magstephalf}
\def \real{\mathbb R}

\def \measure{\rho}

\newcommand{\normp}[2]{||#1||_{#2}}

\def \vspan{\textup{span}}

\newcommand{\innerprod}[2]{\left\langle#1, #2\right\rangle}

\def \samplespace{\Phi}
\def \eventspace{\mathcal F}
\def \probmeasure{\mathcal P}

\def \uniform{\mathcal U}

\def \Prob{\textup{Pr}}
\def \probb{{p}}
\def \distrib{{F}}

\def \qrobb{{q}}

\def \pdf{PDF}
\def \pdfs{f}

\newcommand{\EVs}[1]{\mathbb{E}\left[#1\right]}
\newcommand{\EVsup}[2]{\mathbb{E}_{#1}\left[#2\right]}

\newcommand{\calspace}[1]{{\mathcal#1}}
\def \Hspace{{\mathcal H}}

\ifdefined\icml
\else
    \newtheorem{theorem}{Theorem}
    
    \newtheorem{lemma}{Lemma}

\fi

\newtheoremstyle{proof_sty1}% hnamei
{5pt}% hSpace abovei
{8pt}% hSpace belowi
{}% hBody fonti
{0em}% hIndent amounti
{\scshape}% hTheorem head fonti
{.}% hPunctuation after theorem headi
{.5em}% hSpace after theorem heading
{}% hTheorem head spec (can be left empty, meaning `normal')i
\theoremstyle{proof_sty1}
\newtheorem*{myproof}{Proof}
\def \qed {\hfill \vrule height6pt width 6pt depth 0pt}

\theoremstyle{empty}

\newcommand{\refthm}[1]{Theorem~\ref{#1}}
\newcommand{\reflemma}[1]{Lemma~\ref{#1}}

\theoremstyle{example}
\newtheorem{example}{Example}
\newtheorem{nonexample}{Non-Example}

\def \data{\mathcal D}

\newcommand{\remove}[1]{}

\newcommand{\refeqn}[1]{(\ref{#1})}  % IEEE Computer Society Publication standards
\newcommand{\tabincell}[2]{\begin{tabular}{@{}#1@{}}#2\end{tabular}}
\newcommand{\reffigure}[1]{Figure~\ref{#1}}  % IEEE Computer Society Publication standards
\newcommand{\reftable}[1]{Table~\ref{#1}}
\newcommand{\refsection}[1]{Section~\ref{#1}}

\newcommand{\refapp}[1]{Appendix~\ref{#1}}
\newcommand{\refalg}[1]{Algorithm~\ref{#1}}

\def \eg{{e.g.}}

\def \ie{{i.e.}}
\def \iid{{i.i.d. }}

\def \etc{{etc.}}
\def \bs{\boldsymbol}
\def \vs{{vs.}}
\def \perse{\textit{per se}}
\def \interse{\textit{inter se}}

\def \ipso{\textit{ipso facto}}

\newcommand{\PreserveBackslash}[1]{\let\temp=\\#1\let\\=\temp}
\newcolumntype{C}[1]{>{\PreserveBackslash\centering}p{#1}}
\newcolumntype{R}[1]{>{\PreserveBackslash\raggedleft}p{#1}}
\newcolumntype{L}[1]{>{\PreserveBackslash\raggedright}p{#1}}
\makeatletter
\newcommand\figcaption{\def\@captype{figure}\caption}
\newcommand\tabcaption{\def\@captype{table}\caption}
\makeatother

\newcolumntype{M}{>{\centering\arraybackslash}m{\dimexpr.25\linewidth-2\tabcolsep}}

\definecolor{mygray}{gray}{.93}

\def \figuredir{figure}

\def \var{\textup{Var}}

\def \Impact{\mathscr I}

\def\metricD{{\mathtt d}}

\def \sspace {\calspace{X}}

\def \action{{a}}
\def \utility{{u}}
\def \Utility{{U}}

\def \state{{s}}
\def \staterv{S}
\def \statespace{\calspace{S}}

\def \path{{\bs P}}

\def \unitvec{n}
\def \EU{EU}

\def \figext{pdf}

\def \figcapmaker{\textbf}

\SetAlgoCaptionSeparator{}
\SetAlCapNameFnt{\scshape}
\icmltitlerunning{
Transcendental Idealism of Planner:
Evaluating Perception from Planning Perspective for Autonomous Driving
}

\begin{document}

\twocolumn[ \icmltitle{
Transcendental Idealism of Planner:\\
Evaluating Perception from Planning Perspective for Autonomous Driving
}

\begin{icmlauthorlist}
\icmlauthor{Wei-Xin Li}{q}
\icmlauthor{Xiaodong Yang}{q}
\end{icmlauthorlist}

\icmlaffiliation{q}{QCraft, Santa Clara, CA 95054, USA}

\icmlcorrespondingauthor{Xiaodong Yang}{xiaodong@qcraft.ai}
\icmlkeywords{Autonomous Driving, Planning, Transcendental Idealism}

\vskip 0.3in]

\printAffiliationsAndNotice{} 

\begin{abstract}
Evaluating the performance of perception modules in autonomous driving is one of
the most critical tasks in developing the complex intelligent system. While
module-level unit test metrics adopted from traditional computer vision
tasks are feasible to some extent, it remains far less explored to
measure the impact of perceptual noise on the driving quality of
autonomous vehicles in a consistent and holistic manner.
In this work, we propose a principled framework that
provides a coherent and systematic understanding of the impact an error in the
perception module imposes on an autonomous agent's planning that actually
controls the vehicle. Specifically, the planning process is formulated as
expected utility maximisation, where all input signals from upstream modules
jointly provide a world state description, and the planner strives for the
optimal action by maximising the expected
utility determined by both world states and actions. We show that, under
practical conditions, the objective function can be represented as an inner
product between the world state description and the utility function in a
Hilbert space. This geometric interpretation enables a novel way to analyse the
impact of noise in world state estimation on planning and leads to a universal
 metric for evaluating perception. The whole framework resembles the idea of
transcendental idealism in the classical philosophical literature, which gives the
name to our approach.
\end{abstract}

\section{Introduction}
Autonomous driving has recently emerged as a rapidly advancing realm in both
academia and industry, attracting a surge of interest from scientific and engineering
communities.
As a complex system, an autonomous vehicle (AV) comprises numerous hardware
components and interactive onboard modules.
One such core component is the onboard perception module~\citep{Feng2022},
which serves as the major source of real-time characterisation of the dynamic and
stochastic environment an AV navigates through.

To evaluate and improve the perception module, conventional perception
tasks~(such as detection and tracking) have been well-defined and established with
corresponding performance measurements in computer vision for
benchmarking~\citep{coco, nuscenes}. Despite
their great success in advancing perceptual
information processing modules, almost all such metrics exclusively focus on the
perception-centric performance in a {\it deployment-agnostic} fashion,
ignoring the actual impact of the result to the entire AV system.
Indeed, not all perception errors render the same consequence on AV planning:
missing an obstacle in front of an AV moving forward is obviously far
more serious than one behind.
This problem is further compounded by the
heterogeneity of perceptual errors that share few semantics in common (`How
does an error of 5m/s in velocity compare to that of a size 25\%
larger?'), where manual engineering based on intuition is widely
adopted~\citep{nuscenes,deng2021revisiting}. Although these issues are typically
addressed through integration road tests in the real world, the process is extremely costly and
time-consuming, if not infeasible~\citep{Wachenfeld2016, 2017sljung}.
Consequently, tools are in great demand to effectively and efficiently measure
the performance of perception in the context of the entire AV
system before test or deployment on the road. Unfortunately, these solutions still
remain largely unexplored in the literature.

%%%%%%%%%%%%%%%%%%%%%%%%%%%%%%
\begin{figure*}[t]
		\centering
		\includegraphics[width=0.84\linewidth]{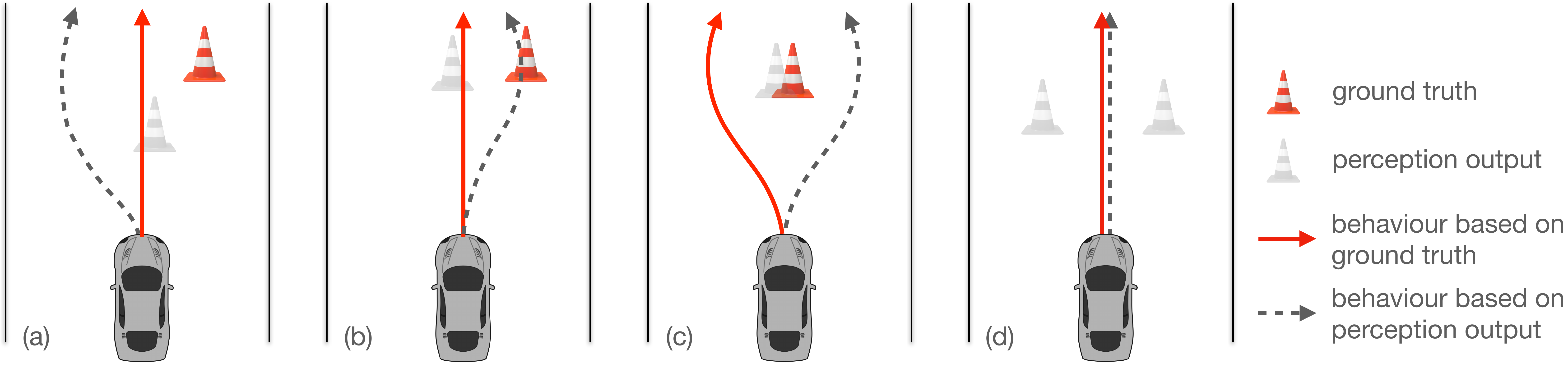}
		\caption{\label{fig:behaviour}\figcapmaker{Illustration of behaviour change
		versus driving cost}. The change in AV behaviour due to a
		perception error is not always correlated to the cost of consequence. In (a)
		the AV has to circumvent the erroneously perceived cone by making a large
		detour. While in (b) the AV only needs to make a slight detour to the
		right, yet it inevitably hits the cone. In this case, although the behaviour
		change is far less than that of (a), the consequence is significantly
		worse~(`hitting an object' versus `making a large detour'). In (c) the
		consequence of either way is indifferent to the AV moving forward, yet
		the change in behaviour is considerable in terms of the spatiotemporal motion.	
		In (d), though two falsely detected cones are
		close to the AV on both sides when passing by without collision, the AV still
		decides to maintain the same motion as in the ground truth case. In this scenario,
		the AV's final behaviour remains the same regardless of the perception error,
		yet the cost of passing by two close obstacles already changes the planning
		process, which cannot be captured by the metrics that only look at the AV
		behaviour or planning result.
  }
\vspace{-3mm}
\end{figure*}
%%%%%%%%%%%%%%%%%%%%%%%%%%%%%%
Recently, the community has begun to approach this problem with some initial
efforts~\citep{waymo-open,Philion2020,deng2021revisiting,Ivanovic2021}.
Despite encouraging results, these preliminary solutions only address certain aspects
of the problem, either implicitly relying on weak correlation between behaviour
change and driving cost~\citep{Philion2020}, inferring the holistic cost
via local properties~\citep{Ivanovic2021},
 or at coarse levels~\citep{waymo-open}. In this work, we propose a principled
and universal framework to quantify how noise in perception input affects AV
planning. This is achieved by explicitly analysing the planning process in the
context of expected utility maximisation~\citep{osborne1994course}, and
evaluating the change in the utility function critical to the AV reasoning subject to
input perception errors.
We show that, under some practical conditions (\refsection{sec:hilbert}),
the planning process can be formulated as an optimisation problem
with a linear objective function in a Hilbert space, where the utility to
optimise is the inner product of an action-wise utility function and the world
state distribution represented by perception. This geometric interpretation
reveals many natural and insightful properties of the problem.
For instance, any
input error can be decomposed into two components: one does not affect the
utility comparison (\textbf{planning-invariant error}), and the other one
directly changes the planning problem (\textbf{planning-critical error}).
Based on this novel insight, we derive a metric to quantify the consequence of
a perception error in changing the planning process.

We want to emphasise the necessity of understanding the impacts of perception errors
on an autonomous driving system via the planning process, rather than solely
from the final result~(\ie~the AV behaviour, or the trajectory output from the
planner), as proposed by previous works~\citep{Philion2020}. This
results from the fact that the final planning result does not necessarily
reflect how AVs evaluate the situation, reason with the environment, and assess
the costs of actions. In fact, the correlation between behaviour change and the
actual consequence is weak, or even negative in many common cases, as
illustrated in~\reffigure{fig:behaviour}. In addition, most works implicitly or
explicitly integrate {\it a priori} knowledge of the consequences of perception errors
into the metric design. The complexity of such impact on autonomous driving,
however, is far beyond handcrafted rules, defeating their purposes despite
tremendous amounts of manual efforts. For instance, \citet{deng2021revisiting} assume
that the severity of an error should be weighted proportional to the reciprocal of
its cubed Manhattan distance to the AV, regardless of its position relative to
the latter.
These presumptions, without convincing justification, could introduce
 biases that conceal crucial facts for evaluation purposes~(see \refsection{sec.exp.sync.case}
for an example).
In contrast, we make few such assumptions and solely rely on the planning
process to infer the error consequence in a completely unbiased fashion, which enables
our solution to capture many critical or subtle cases.
In this regard, the core principle of our design resembles the
philosophical concept of {\it transcendental idealism} proposed by Immanuel Kant
in his classical work {\it Critique of Pure Reason}~\citep{kant_1781}, which
argues that, due to the limitation of the observer's sensibility, the cognition of
external objects is processed never as they are \perse, but via the
cognitive faculties and subject to the interpretation of the observer's
experience. For the same reason, the properties~(\eg~ignorability, impact) of
a perception error (an external object) \ipso~should be understood through
the corresponding disturbance it causes to the AV planner (the
 observer)
 and measured by the extra loss incurred from the planning viewpoint,
 which gives the name to our framework:
\textbf{transcendental idealism of planner~(TIP)}. 
Our code is available at \url{https://github.com/qcraftai/tip}. 

\vspace{-2mm}
\section{Literature Review}
\label{sec:literature}
\vspace{-1mm}
{\bf Metrics for AV Perception Evaluation.}
Recent works aimed to assess the performance of perception from the autonomous
 driving system viewpoint mostly approach the problem in heuristic ways.
 Multiple heterogeneous detection metrics are directly combined to produce a single score
 for detector evaluation in the popular nuScenes benchmark~\citep{nuscenes}.
 Considering neural planners, \citet{Philion2020} implicitly
 hypothesise that consequences of perception errors on driving are directly
 correlated to the change in the planned spatiotemporal trajectories of an AV, and
 propose the planning KL-divergence~(PKL) to measure the impact.
While intuitive, it fails to incorporate the context of the environment and does not
 precisely reflect the real cost of perception noises in many common traffic
 scenarios.
To address the specific problem of object representation,
\citet{deng2021revisiting} study how object shapes can affect autonomous driving
and devise the support distance error~(SDE) to quantify the effect.
In another recent work, \citet{Ivanovic2021} look into the planning process and
employ sensitivity as a probe of the input signal's contribution to AV behaviour.
This, however, only leverages local properties of differentiable cost
functions to infer global results.
In comparison, our approach systematically captures the global properties
of the planning process and applies to more general cases.
For convenience, the comparison of these metrics is summarised in~\refapp{sec:more}.

\vspace{-1mm}
{\bf Planning for Autonomous Vehicles.}
In this work, we consider both behavioural decision making and motion planning as
{\it the planning process}, which generates the vehicle behaviour
 for the controller to execute given the observation up to the planning time.
There is a rich literature to address this fundamental problem for AVs, 
which can be roughly categorised into {\it utility-based} and
{\it utility-free} methods.
The former typically relies on a utility function to encode the predefined goals 
and strives for the optimal behaviour to accomplish the
maximum return, 
via solving an optimisation problem with an explicit cost function manually
 engineered or learnt to guide vehicle trajectory 
 generation~\citep{Buehler2009, Werling2010, planner, fan2018baidu, Ajanovic2018}, 
searching for the action policy with best reward return in the reinforcement learning 
framework~\citep{Kuefler2017,Schwarting2018, Kendall2019, Bronstein2022}, 
\etc~ 
The latter learns to drive by directly mapping input signals
(raw or processed sensor data) into AV behaviours or vehicle
control commands by leveraging
deep learning from
massive data~\citep{bojarski2016,Guez2019,Grigorescu2020} as an alternative, which has attracted
increasing attention from the research community recently. 
In spite of some
promising results, nontrivial challenges still remain for this paradigm. For
instance, behaviour cloning~\cite{Muller2005,Bansal2019,Prakash2021}, one of the
most popular strategies along this line, seeks to approach the human driving
capability by learning from a large corpus of driving records available from
human daily driving activities. Besides the considerable demand for supervised driving
experiences to cover as many rare situations as possible for reliability, it
 suffers from generalisation issues by domain shift between training and
deployment~\citep{Codevilla2019, Haan2019}, as well as the inability, due to
its open-loop learning nature, to infer long-term interaction between the AV
and the environment~\citep{Zhang2022}, a critical merit for handling complex
traffic situations.
In this work, we aim to exploit the properties of AV planners with
 explicit rewarding mechanisms to shed light on the impacts of
 perception noise on this process, and focus on utility-based
 planning.

\vspace{-2mm}
\section{Planning as Expected Utility Maximisation}
\label{sec:planning}
\vspace{-1mm}
To introduce our approach,
we first present preliminary math basics to facilitate discussion,
 then review the expected utility maximisation as the optimal AV action
framework, followed by its interpretation in a Hilbert space, based on which
our metric for perception evaluation is derived.

\vspace{-2mm}
\subsection{Preliminaries}
\label{sec:preliminary}
Unless otherwise specified explicitly, all notation follows the
standardised one in~\citep{Goodfellow-et-al-2016}.
A probability space $\{\samplespace, \eventspace, \probmeasure\}$ is defined by
a sample space $\samplespace$, an event space $\eventspace$ (a $\sigma$-algebra
on $\samplespace$), and a Borel probability measure $\probmeasure$ on
$\eventspace$. A random variable $X: \samplespace\rightarrow\real^d$
($d\in\natural$) is induced from $\{\samplespace, \eventspace, \probmeasure\}$
with distribution function $\distrib_X(x)$. When absolutely continuous,
$\distrib_X(x)=\int_{-\infty}^x\pdfs_X(t)\dif t$, where $\pdfs_X(x)$ is the
probability density function~(\pdf).
$L^2(\sspace, \measure)$ denotes the space of square-integrable functions, and
$\measure$ is a Lebesgue measure accordingly. A Hilbert space
$\Hspace=({\mathcal T}, \innerprod{\cdot}{\cdot})$ is defined on a complete
space ${\mathcal T}$ with inner-product $\innerprod{\cdot}{\cdot}_\Hspace$ and
induced norm $\norm{\cdot}_\Hspace$. Let $S\subset\Hspace$ be a subspace of
 $\Hspace$, $S^\perp=\{x\in\Hspace|\innerprod{x}{y}, \forall y\in
S\}$ is the orthogonal complement of $S$~(\ie~the set of all vectors
orthogonal to $S$). $\vspan(S)$ is the linear span of a set $S$.
$\unitvec_v\coloneqq{v}/{\norm{v}}$ is an element of unit length in a normed
vector space by normalising element $v$.

\vspace{-2mm}
\subsection{Autonomous Vehicles as Rational Agents}

An AV is an intelligent agent that aims to accomplish some predefined goals in
an interactive and uncertain environment. It is constantly faced
with planning problems in the dynamic surroundings, and the quality of
planning determines how well the goals can be achieved. By the classical
expected utility maximisation~(EUM) theory~\citep{osborne1994course},
at any given time $t$, an AV aims to
achieve the maximum expected reward, defined by the utility function $\Utility$,
via execution of the optimal action $\action^*_t$ such that
\begin{equation}
\vspace{-0.5mm}
\action^*_t =
 \argmax\nolimits_{\action\in\data_{\action,t}}~
 \EVs{\Utility(\staterv_t, \action)},
\label{eq:eum}
\end{equation}
where $\data_{\action,t}$ is the set of all feasible AV actions at time $t$;
$\state\in\calspace{S}$ is the state random variable at time $t$ with distribution function
$\distrib_{\staterv_t}(\state)$ in the world
state space $\calspace{S}$; and
\begin{equation}
\EU(\distrib_{\staterv_t}, a)
\coloneqq \EVs{\Utility(\staterv_t, \action)}
= \scalebox{1.3}{$\int$}_{\state\in\calspace{S}}~\Utility(\state, \action) \dif \distrib_{\staterv_t}(\state).
 \nonumber
\end{equation}
Intuitively, the utility function encodes the goal or reward the AV is supposed
to achieve (\eg~reaching a destination in time, avoiding collision with other objects).
$\distrib_{\staterv_t}(\state)$ captures uncertainty about the
stochastic environment given all world knowledge and historical
observations up to $t$, which are estimated by modules like localisation and
perception~\citep{pillar-motion, simtrack}.
Architectures of many modern AV planners still follow the concept of
this classical framework as its
variants~\citep{planner,fan2018baidu,Sadat2020,Bronstein2022,Zhang2022}.

\vspace{-1mm}
\subsection{Expected Utility Maximisation in Hilbert Space}
\label{sec:hilbert}
\vspace{-1mm}
To gain some insights into the expected utility of~\refeqn{eq:eum} and how input
 noises affect the planning process,
we introduce an interpretation in the Hilbert space to leverage geometric tools
available from linear algebra. We first establish the conditions under which a
probability measure can be embedded into a Hilbert space, followed by the
interpretation of EUM from a geometric perspective in~\refsection{sec:metric}.
For brevity, all proofs are left in~\refapp{app:proof}.

\begin{theorem}[Probability Measure Embeddings in the Hilbert Space]
\label{thm.hilbert}
Let $\{\sspace, \metricD\}$ be a compact
 metric space with $\metricD$ as the metric function, $\probb$ be a Borel
probability measure on $\sspace$, and $X$ be a random variable on $\sspace$ with
distribution function $\distrib_X(x)$. If $\distrib_X(x)$ is absolutely
continuous
and the density function $\pdfs_X$ is square-integrable, \ie~$\pdfs_X\in L^2$,
then there exists a unique element\footnote{It is a family of
functions that are equal almost everywhere.}$\mu_\probb\in\Hspace$ such that
\vspace{-1mm}
\begin{equation}
\EVsup{X}{g(x)} =
\innerprod{{\mu_\probb}}{g}_{\Hspace},~\forall g\in\Hspace,
\label{eq:hilbert.embedding}
\vspace{-1mm}
\end{equation}
where element $\mu_\probb$ denotes the {\it embedding} of probability measure
$\probb$ in the Hilbert space $\Hspace=(L^2, \innerprod{\cdot}{\cdot})$, with
the inner product given by
\vspace{-1mm}
\begin{equation}
\innerprod{g}{h}_{\Hspace} \coloneqq \scalebox{1.1}{$\int$}_{x}~g(x)h(x)\measure(\dif x).
\nonumber
\vspace{-1mm}
\end{equation}
\end{theorem}
The critical condition of $\distrib_X(x)$ being absolutely continuous with a
square-integrable density function $f_X$ in \refthm{thm.hilbert} is
general and includes many common distributions as special cases (see~\refapp{app:example}).
The mapping from probability measures of
continuous random variables to $\Hspace$ 
established by \refthm{thm.hilbert} is also {\it
injective}
by the following result.
\begin{theorem}[Injection of Probability Measure Embeddings]
\label{thm.hilbert.injection}
Let $\probb$ and $\qrobb$ be two Borel probability measures defined on a compact
metric space $\{\sspace, \metricD\}$ with absolutely continuous distribution
functions, then $\probb=\qrobb$ almost everywhere if and only if
$\mu_\probb=\mu_\qrobb$, where $\mu_\probb$ and $\mu_\qrobb$ are the embeddings
of $\probb$ and $\qrobb$ in $\Hspace$, respectively.
\end{theorem}

A similar result for mixed distributions is also available in~\refthm{thm.hilbert.mixed}
for deterministic perception
results~(treated as Dirac delta distributions) and
guarantees that following discussion on probabilistic perception results can
be readily extended to these cases.
Under the conditions in the aforementioned results, the EUM of \refeqn{eq:eum} can be rewritten as
\vspace{-1mm}
\begin{equation}
\hspace{-1.5mm}
\action^*
=\argmax\limits_{\action\in\data_{\action}}\EVsup{\probb(\state)}{\Utility(\state,\action)}
=\argmax\limits_{\action\in\data_{\action}}\innerprod{{\mu_\probb}}{\Utility_\action}_{\Hspace}.
\label{eq:eum.hilbert}
\end{equation}
Given this injective correspondence between $\probb$ and $\mu_\probb$,
we can leverage many tools in algebra (\eg~inner product, orthogonality,
projection)
to analyse the impact of perception errors on AV
planning via the EUM in $\Hspace$,
where the topological structure is exclusively determined by its inner product.

%%%%%%%%%%%%%%%%%%%%%%%%%%%%%%%%
\def\ew{1.05in}
\begin{figure}[t]
	\centering
	\includegraphics[width=0.55\linewidth]{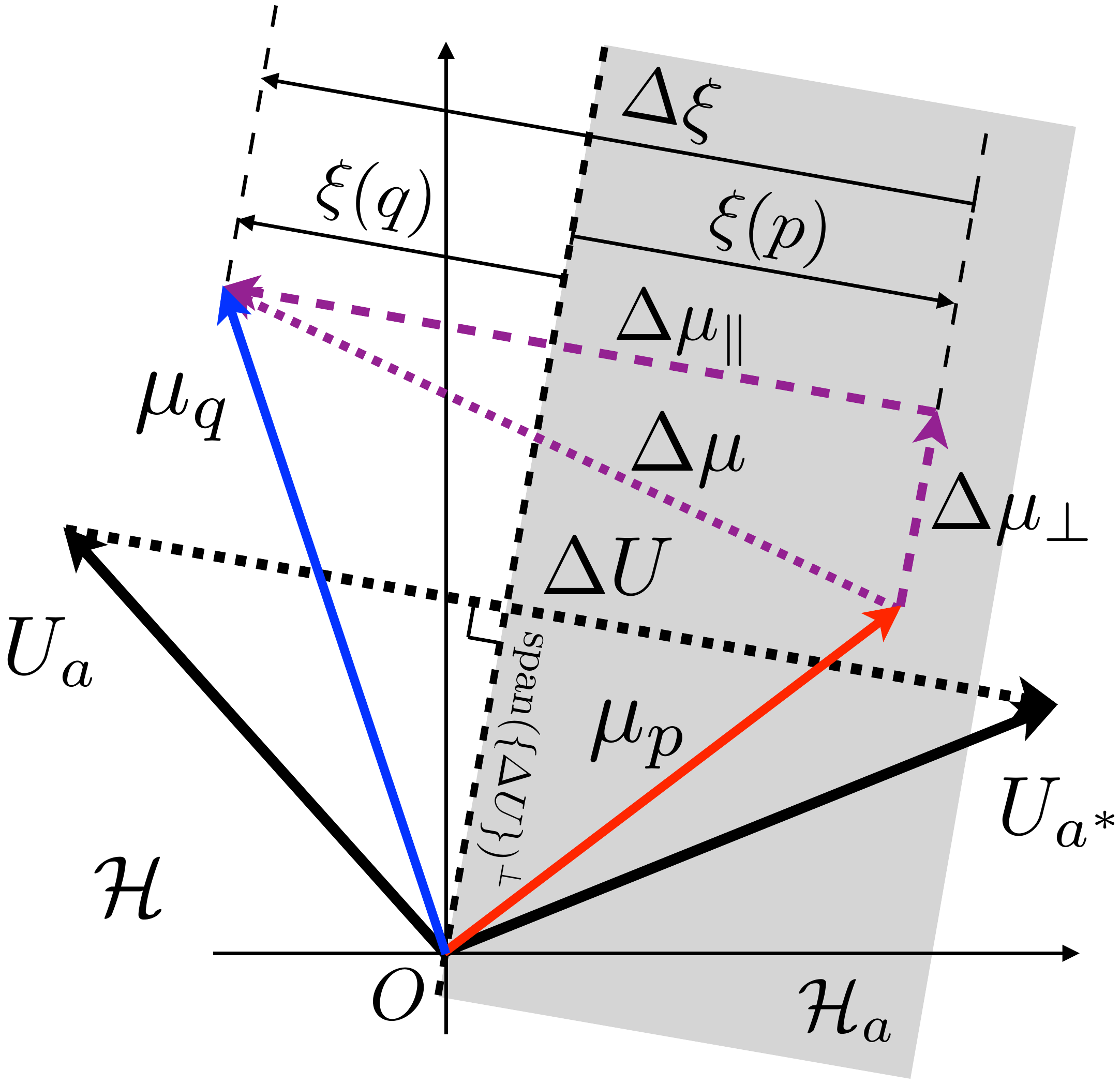}
	\figcaption{\label{fig.H}\figcapmaker{Illustration of EUM in $\Hspace$.}
        $\Delta\Utility=\Utility_{\action^*}-\Utility_{\action}$ defines the
		behaviour direction;
		$\xi$ is the preference score; $\mu_\probb$ and
		$\mu_\qrobb$ are the embeddings of the ground truth and perception results, respectively;
		$\Delta\mu$ is the perception error, which is
		decomposed into the planning-critical error~(PCE) $\Delta\mu_\parallel$, and
		the planning-invariant error~(PIE) $\Delta\mu_\perp$; and the shaded area
		corresponds to $\Hspace_\action$.
		Note that ${\footnotesize\langle{\Delta\mu_\parallel},{\Delta\Utility}\rangle}<0$. }
\vspace{-2mm}
\end{figure}
%%%%%%%%%%%%%%%%%%%%%%%%%%%%%%%%

\vspace{-1mm}
\section{Perception Evaluation via AV Planning}
\label{sec:metric}
\vspace{-1mm}

In this section, we derive the effect of perception
errors on planning via the theoretical foundation established
in~\refsection{sec:planning}. 
Without loss of generality, we assume that the perception module is the only
source for world state estimation in the following discussion. 

\vspace{-1mm}
\subsection{Breakdown of Perception Errors}
\label{sec:error.breakdown}
\vspace{-1mm}

%%%%%%%%%%%%%%%%%%%%%%%%%%%%%%%%
\def\ew{1.05in}
\begin{figure}[t]
 \setlength\parindent{-1mm}
	\centering
	\includegraphics[width=0.95\linewidth]{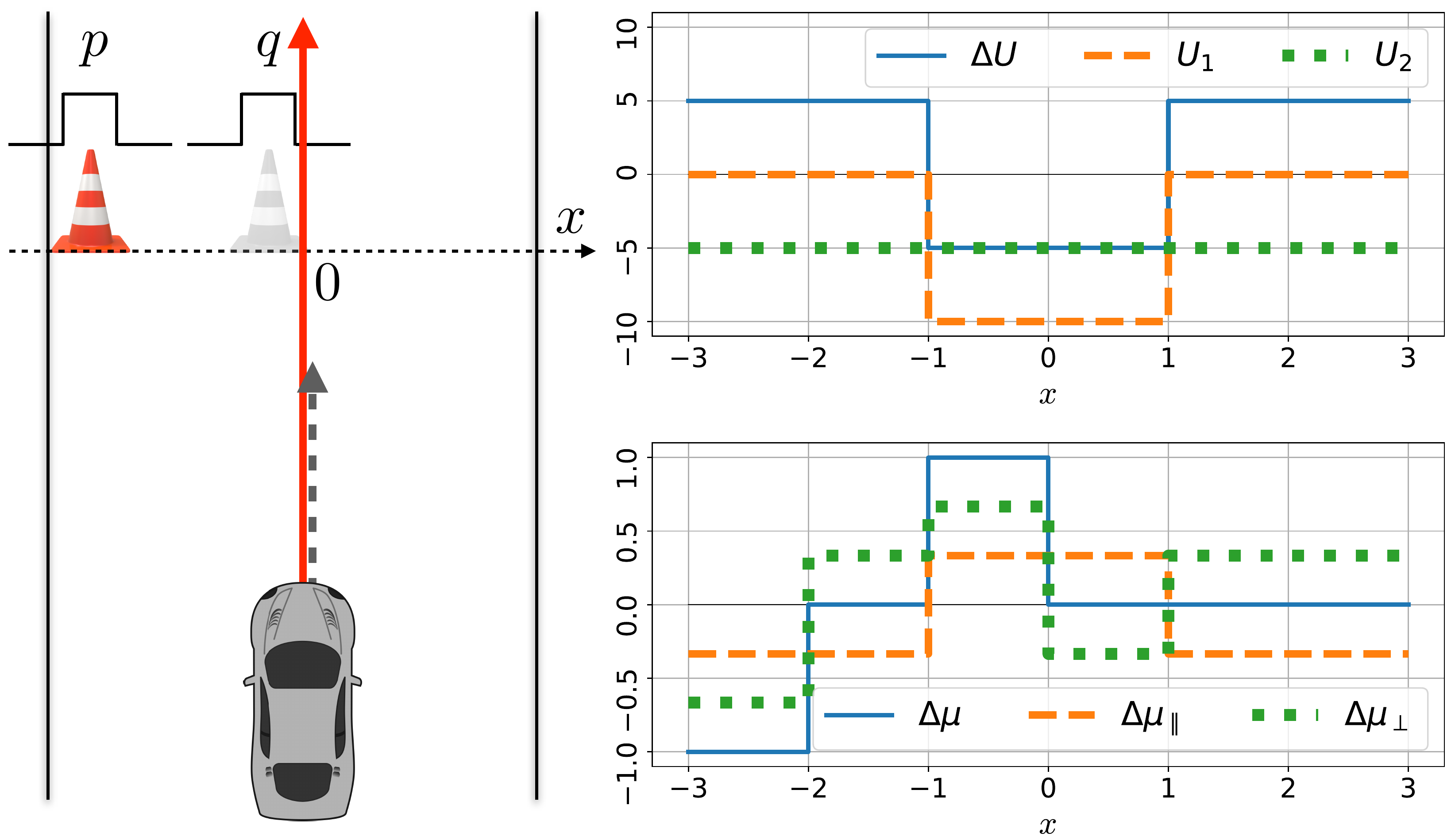}%
  \vspace{-1mm}
	\figcaption{\label{fig.pce.example}\figcapmaker{An example of PCE $\Delta\mu_\parallel$ and PIE $\Delta\mu_\perp$.}
        An AV is moving forward on a $6$m-wide road;
		there is a cone in front on a line across the road (the $x$ axis).
		The ground truth distribution of the cone location
		$\probb$ is $\uniform_{[-3,-2]}$, a uniform distribution with support
		$[-3,-2]$, while the perception predicts its location distribution
		$\qrobb$ to be $\uniform_{[-1,0]}$. The 2m-wide AV has two action options:
		(\romannumeral 1) to keep
		moving forward ($\action^*$, the solid line with an arrowhead), and the utility function is
		$\Utility_1(x) = -10\cdot\1{x\in[-1, 1]}$ with $x$ being the position of the
		cone (only large loss for collision with the cone);
		(\romannumeral 2) to come to a full stop before the line via hard braking
		($\action$, the dashed line with an arrowhead), and the utility function is a constant
		$\Utility_2(x)=-5$ (loss of hard braking is consistent regardless of the cone
		position).
		Note that $\Delta\mu_\parallel$ is of the same shape as
		$\Delta\Utility=\Utility_1-\Utility_2$ (up to a negative constant), and
		$\innerprod{\Delta\Utility}{\Delta\mu_\perp}_{\Hspace}=0$.
		In this example,  PCE accounts for 33.3\% of the error energy, while PIE for 66.6\%.
		See \reffigure{fig.pbe.example} for another case where ${\footnotesize\innerprod{\Delta\mu}{\Delta\Utility}_{\Hspace}>0}$.}
\vspace{-2mm}
\end{figure}
%%%%%%%%%%%%%%%%%%%%%%%%%%%%%%%%

Consider a general case where the candidate action set is
$\data_\action=\{\action_i\}$, and each action is associated with a distinct
utility function $\Utility(\state,\action_i)\in\Hspace$ such that, $\forall\action_i, \action_j\in\data_\action$,
$$\norm{\Utility(\state,\action_i)-\Utility(\state,\action_j)}_\Hspace > 0
\Leftrightarrow
\action_i\neq\action_j.
$$
Let
$\action^*$ be the optimal action per EUM of~\refeqn{eq:eum.hilbert} given the
ground truth world state distribution $\probb(\state)$. For any
$\action\neq\action^*$, $\Delta\Utility({\action^*},
\action)=\Utility_{\action^*}-\Utility_{\action}$; the planning half-space
in $\Hspace$ is
\begin{equation}
\Hspace_{\action}\coloneqq\{f|\innerprod{f}{\Delta\Utility({\action^*}, \action)}_{\Hspace}>0, f\in\Hspace\}.
\nonumber
\end{equation}
Given the perception result $\qrobb(\state)$, $\action^*$ is preferred
over $\action$ by EUM if and only if $\mu_\qrobb\in\Hspace_{\action}$, \ie
\begin{equation}
\label{eq:action.preference}
\xi(\qrobb; \action^*, \action) > 0
 \vspace{-2mm}
\end{equation}
with
\vspace{-1mm}
\begin{equation}
\label{eq:action.preference.def}
\hspace{-2mm}
\xi(\qrobb; \alpha, \beta)
\coloneqq\innerprod{{\mu_\qrobb}}{\Delta\Utility({\alpha}, \beta)}
=\EU(\qrobb, \alpha) - \EU(\qrobb, \beta)
\end{equation}
denoting the $\alpha$-$\beta$ preference score
given $\qrobb$~($\forall\alpha, \beta\in\data_\action$),
which exclusively decides the result of EUM. As illustrated
in~\reffigure{fig.H}, the planning result remains $\action^*$ if and only if
\vspace{-1mm}
\begin{equation}
\mu_\qrobb\in\scalebox{1}{$\bigcap\nolimits_{\action\in\data_\action/\{\action^*\}}$}\Hspace_{\action}.
\nonumber
\end{equation}
When $\qrobb(\state)$ is erroneous (\ie~$\norm{\mu_\qrobb-\mu_\probb}_\Hspace>0$),
the preference score of \refeqn{eq:action.preference} may be affected, \ie~$\xi(\qrobb; \action^*,
\action)\neq\xi(\probb; \action^*, \action)$, so is the result by EUM.
To understand how error
$\Delta\mu=\mu_\qrobb-\mu_\probb$ changes the result of EUM, we further
decompose $\Delta\mu$ into two orthogonal components:
\vspace{-1.5mm}
\begin{equation}
\Delta\mu
= \mu_\qrobb-\mu_\probb
= {\Delta\mu}_\parallel +{\Delta\mu}_\perp,
\vspace{-2mm}
\end{equation}
where
\vspace{-2mm}
\begin{equation}
{\Delta\mu}_\parallel
= \innerprod{\Delta\mu}{\unitvec_{\Delta\Utility}}_{\Hspace} \unitvec_{\Delta\Utility}
= \frac{\innerprod{\Delta\mu}{\Delta\Utility}_{\Hspace}}{\norm{\Delta\Utility}_\Hspace^2}\Delta\Utility
\end{equation}
is the projection of $\Delta\mu$ onto unit vector
$\unitvec_{\Delta\Utility}$~(denoted \textit{behaviour direction}),
and
${\Delta\mu}_\perp\in\vspan(\{{\Delta\Utility}\})^\perp$
is the projection of ${\Delta\mu}$ onto the orthogonal complement of the
subspace spanned by the behaviour direction, \ie~$\innerprod{\Delta\mu_\perp}{\Delta\Utility}_{\Hspace}=0$.
In the presence of error $\Delta\mu$, as illustrated
in~\reffigure{fig.H} and~\reffigure{fig.H.pbe}, the change in preference score of
\refeqn{eq:action.preference} is only determined by ${\Delta\mu}_\parallel$:
\begin{align}
\label{eq:decomp}
\Delta\xi(\action^*, \action; \qrobb, \probb)
=&~\xi(\qrobb; \action^*, \action) - \xi(\probb; \action^*, \action)\\
=&~{\footnotesize\innerprod{\Delta\mu}{\Delta\Utility}_{\Hspace}} \nonumber\\
=&~{\footnotesize\langle{\Delta\mu_\parallel},{\Delta\Utility}\rangle}_{\Hspace}. \nonumber
\end{align}
For this reason, we denote $\Delta\mu_\parallel$ as the \textbf{planning-critical
error~(PCE)},
and $\Delta\mu_\perp$ as the \textbf{planning-invariant error~(PIE)}~(see
\reffigure{fig.pce.example} and~\refapp{sec:pce} for more discussion).
The observation reveals two pivotal facts: {\it (\romannumeral 1)~not all errors
in perception
(world state estimation)
are of equivalent impact on planning, and the
subspace $\vspan(\{{\Delta\Utility}\})^\perp$ contains all errors that do not
affect EUM at all; (\romannumeral 2) errors in subspace $\vspan(\{{\Delta\Utility}\})$
either negatively affect planning~(if
${\footnotesize\innerprod{\Delta\mu}{\Delta\Utility}<0}$), or even favour the optimal
action $\action^*$~(if ${\footnotesize\innerprod{\Delta\mu}{\Delta\Utility}>0}$).}
Intuitively, $\Delta\xi$ measures the impact of perception error $\Delta\mu$ on
 the decision between $\action^*$ and $\action$.

\vspace{-1mm}
\subsection{Estimation of Preference Score $\xi$}
\label{sec:eu}
In practice, combining \refeqn{eq:action.preference.def} and \refeqn{eq:decomp},
evaluating the impact of a perception error $\Delta\mu$ on $\action^*$-$\action$ decision is reduced to
\vspace{-2mm}
\begin{equation}
\begin{split}
\Delta\xi(\action^*,~&\action;~\qrobb,~\probb) \\
=~&\EVsup{\qrobb(\state)}{\Utility(\state, \action^*)} - \EVsup{\probb(\state)}{\Utility(\state, \action^*)}\\
~&-\EVsup{\qrobb(\state)}{\Utility(\state, \action)} + \EVsup{\probb(\state)}{\Utility(\state, \action)}.
\end{split}
\end{equation}

Computing these expectations in analytical forms typically requires strong
 assumptions on the forms of both utility and distribution functions for precise results,
 or variational methods for approximation~\citep{PRML},
 which limits representation capacity or accuracy.
For maximum flexibility,
we resort to numerical methods
by estimating the expected utilities from finite-size samples of world states,
and show that the solution is both statistically consistent and uniformly
efficient under practical conditions.
Specifically, for a fixed action $\action$, given an \iid sample of the
utilities $\{U(\staterv_i, \action)\}_{i=1}^{n}$ with $\staterv_i$ drawn from
$\probb_\staterv(\state)$, an unbiased estimator of the
expected utility via U-statistics is
\begin{equation}
\label{eq:estimator.eu}
EU_\action = \frac{1}{n}\sum\nolimits_{i=1}^n U(\staterv_i, \action).
\end{equation}
A fast convergence rate via the
 uniform bound can be achieved
 by the following observation for the estimator. 
\begin{theorem}[Exponential Convergence Rate]
\label{thm.estimator.bound.exp}
If there exists an $M\in\real$ such that $\abs{\Utility(\staterv,
\action)}<M$ almost surely,
then for $EU_\action$ of~\refeqn{eq:estimator.eu}, $\forall\varepsilon>0$,
\begin{equation}
\label{eq:estimator.eu.hoeffding}
\Prob\Big(\abs{EU_\action-\EVs{\Utility(\staterv, \action)}}>\varepsilon\Big)
< 2\exp\Big(-\frac{n\varepsilon^2}{2L}\Big),
\end{equation}
where $L=\min\Big(M^2, \var\left(\Utility(\staterv,
\action)\right)+\frac{M\varepsilon}{3}\Big)$.
\end{theorem}
Note that, the condition of~\refthm{thm.estimator.bound.exp},
assigning finite values to the utility in the worst or best cases,
is a practical necessity even for life-related
scenarios~\citep[Chapter 16.3]{AI}.
The exponential convergence rate of $O(e^{-n})$ provided by
\refthm{thm.estimator.bound.exp} is significant:
it depends on
(\romannumeral 1) neither the dimensionality of the original state space
$\statespace$, \ie~the curse of dimensionality is not invoked~\citep{AoS}, nor
(\romannumeral 2) the distribution 
and utility functions, \ie~$\Utility(\staterv, \action)$ and $\probb_\staterv(\state)$ can
take any arbitrary forms.

%%%%%%%%%%%%%%%%%%%%%%%%%%%%%%%%%%
\setlength{\textfloatsep}{0.2in}
\setlength{\intextsep}{0pt}
\begin{algorithm2e}[t]
\caption{TIP Score Computation}
\label{alg.tip}

\SetKwInOut{Input}{Input}\SetKwInOut{Output}{Output}

\Input{A query perception sequence
${\qrobb(\{\state_t\}_{t=-\tau}^0)}$, the ground truth
${\probb(\{\state_t\}_{t=-\tau}^0)}$, sample size $n$
}
\Output{TIP score $\Impact(\qrobb, \probb; \Utility, \data_\action)$}

\BlankLine

Get the candidate action set $\data_{\action,\probb}$ and the optimal action $\action^*\in\data_{\action,\probb}$ from the planner with the ground
truth
$\probb$

Get the candidate action set $\data_{\action,\qrobb}$ from the planner with
the query perception input
$\qrobb$

$\data_\action \leftarrow \data_{\action,\probb}\cup\data_{\action,\qrobb}$

\ForEach{$\action\in\data_\action$}{
	$\{\state_{\probb}^{(i)}\}_{i=1}^n$ $\leftarrow$
	$n$ \iid observations from $\probb$

	$\{\state_{\qrobb}^{(i)}\}_{i=1}^n$ $\leftarrow$ $n$ \iid observations from
	$\qrobb$
	 {\small
	\vspace{-3mm}
	 \begin{flalign}
        {\Delta\xi}_\action \leftarrow \frac{1}{n} \sum\nolimits_{i=1}^n &~\Big[\Utility(\state_{\qrobb}^{(i)}, \action^*)
	 		- \Utility(\state_{\qrobb}^{(i)}, \action) &\nonumber \\
                    &~~~~- \Utility(\state_{\probb}^{(i)}, \action^*)
	 		+ \Utility(\state_{\probb}^{(i)}, \action)\Big]& \nonumber
        \end{flalign}
	 \vspace{-7mm}
	 }
}
{$
\Impact(\qrobb, \probb; \Utility, \data_\action) \leftarrow
\min\nolimits_{\action\in\data_{\action}}
{\Delta\xi}_\action
$}

\BlankLine
\end{algorithm2e}
%%%%%%%%%%%%%%%%%%%%%%%%%%%%%%%%%%

\vspace{-2mm}
\subsection{Perception Error Impact on Planning by TIP}
 \vspace{-1mm}
Given the consequence of a perception error evaluated on an action $\action$ in  \refeqn{eq:decomp} ,
its impact on AV planning is defined as
the maximum reduction of preference scores among all candidate actions in $\data_\action$:
\vspace{-2mm}
\begin{equation}
\label{eq:metric}
\Impact(\qrobb, \probb; \Utility, \data_{\action}) \coloneqq
\min\nolimits_{\action\in\data_{\action}}\Delta\xi(\action^*, \action; \qrobb, \probb) \leqslant 0.
\vspace{-2mm}
\end{equation}
This
 leads to the optimal sensitivity to the worst case.
Other alternatives, nevertheless, are also possible for
 different trade-offs between selectivity and invariance, \eg~means,
 top $k$ percentiles~\citep{li2015multiple}.
In our case, the action set contains spatiotemporal
trajectories the planner considers in all phrases during the whole
planning process~(see~\refapp{sec:planner} for more details).
To facilitate the understanding of our approach, the
pseudocode is provided in~\refalg{alg.tip}, which sketches
the basic routine to compute the TIP score of a perception input sequence
$\qrobb(\{\state_t\}_{t=-\tau}^0)$ from $t=-\tau$ to $t=0$ for planning at
$t=0$.

It should be noted that, once the planner utility function is established for
scenario-independent deployment, TIP score evaluation is a parameter-free process,
and results are readily comparable across different scenarios.
This advantage is in contrast to other handcrafted metrics like NDS or SDE-APD,
which require either manual specification~\citep{nuscenes} or calibration~\citep{deng2021revisiting},
thus may behave inconsistently inside and outside the intended noise dynamic range, as will be seen in~\refsection{sec.exp.sync.noise}.

\vspace{-2mm}
\section{Empirical Study}
\label{sec:empirical}
\vspace{-1mm}
In this section, we evaluate how TIP works empirically via
extensive qualitative and quantitative experiments conducted on
both synthetic and real data.

\vspace{-2mm}
\subsection{Basic Settings}
\label{sec:settings}
\vspace{-1mm}
All AVs used in the experiments are based on the same type of regular passenger
vehicles. 
The planner deployed on the AVs
consists of various sub-modules of routing, object motion forecasting, cost
generation, path finder, and trajectory optimisation. At each planning time, 
these sub-modules
 analyse the
environment and input history to establish the target utility function
$\Utility(\cdot, \state)$ for final trajectory optimisation.
The path finder then provides multiple initial paths as
candidates for path-wise trajectory optimisation, and the final choice is
determined by a utility decider.
The goals the planner strives to achieve include motion smoothness,
traffic rule compliance,
safety,
progress to the destination, \etc~The final output trajectory is subject to 
independent Gaussian noise at each time step to account for control inaccuracy.
The planner has been extensively verified via
rigorous road tests in
major cities with millions of population (see~\refapp{sec:planner} for more
details).
Note that, while we adopt a planner following the popular module-based architecture to
evaluate the proposed metric in this work, it can also be readily applied to other utility-based
alternatives, \eg~planners learnt via the
imitation learning~\citep{Kuefler2017}, Markov decision process~\cite{Zhang2022},
or with trajectory density modelling~\citep{Bronstein2022}.

All experiments are implemented in scenarios as the standard protocol in
autonomous driving~\citep{Riedmaier2020}. The scenarios used are collected from real
world road tests~(more details in~\refapp{app:data}).
We consider the planning problem at a particular frame in a scenario at a time,
and evaluate the utility of an action (a spatiotemporal trajectory the AV
executes) for the next three seconds, following the basic setup
of~\citep{Philion2020}.
For comparison, three baselines are adopted from the spectrum of perception
metrics: (\romannumeral1)~at the conventional end, nuScenes dataset
score~(NDS) combines several traditional scoring results for 3D
object detection into a single performance measure~\citep{nuscenes},
(\romannumeral2)~SDE average precision
distance-weighted~(SDE-APD) focuses more on detections
near the AV in an ego-centric fashion~\citep{deng2021revisiting}, and
(\romannumeral3)~PKL~\citep{Philion2020} serves as the representative of AV
behaviour-based metrics in the literature\footnote{
PKL is always nonnegative by definition. 
%and a smaller PKL score indicates a better detection performance.
For ease of comparison with other metrics, in this work 
we follow the practice of \citep{Philion2020} and negate the raw PKL scores.   
}.

\vspace{-2mm}
\subsection{Results on Synthetic Data}
\label{sec:exp.synthetic}
\vspace{-1mm}
In the first set of experiments, we aim to gain some understanding of various
metrics in reaction to common types of perception noises. A dataset is
synthesised from our curated road test scenarios by adding controlled noise to
the 3D object ground truth of vehicles, to enable clear observation of the
sensitivity of metrics to specific perception error types. For this, 1000
5s-long scenarios are assembled, with the number of objects per scene between 30
and 500 and an average AV speed of 5m/s or higher. The ground
truth is annotated by professionally trained human operators. All objects in the
scenario are labelled with location, heading, category, and bounding box from 3D
point clouds recorded by onboard LiDAR sensors during road tests.
%%%%%%%%%%%%%%%%%%%%%%%%%%%%%%%%
\def\ew{1.1in}
\begin{figure}[t]
	\centering
	\includegraphics[width=\ew]{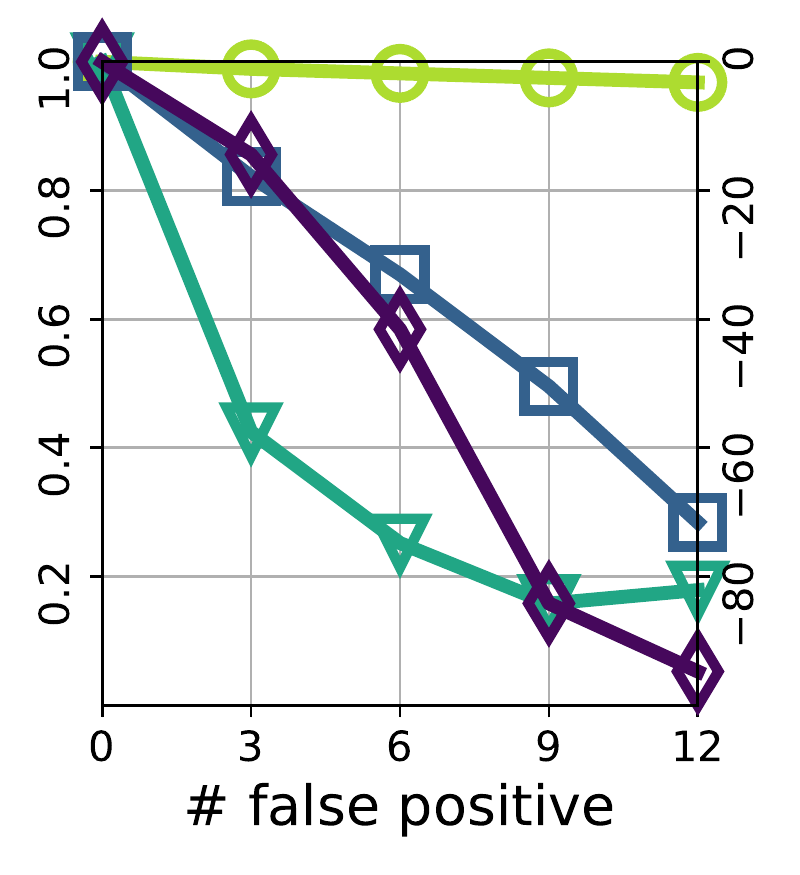}%
        \includegraphics[width=\ew]{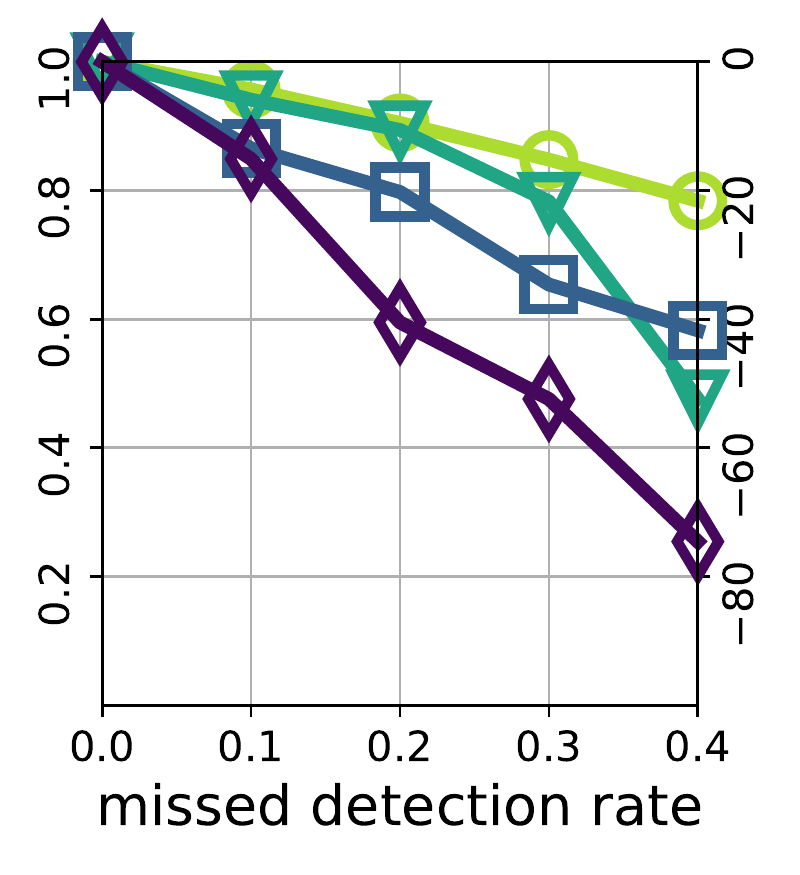}%
        \includegraphics[width=\ew]{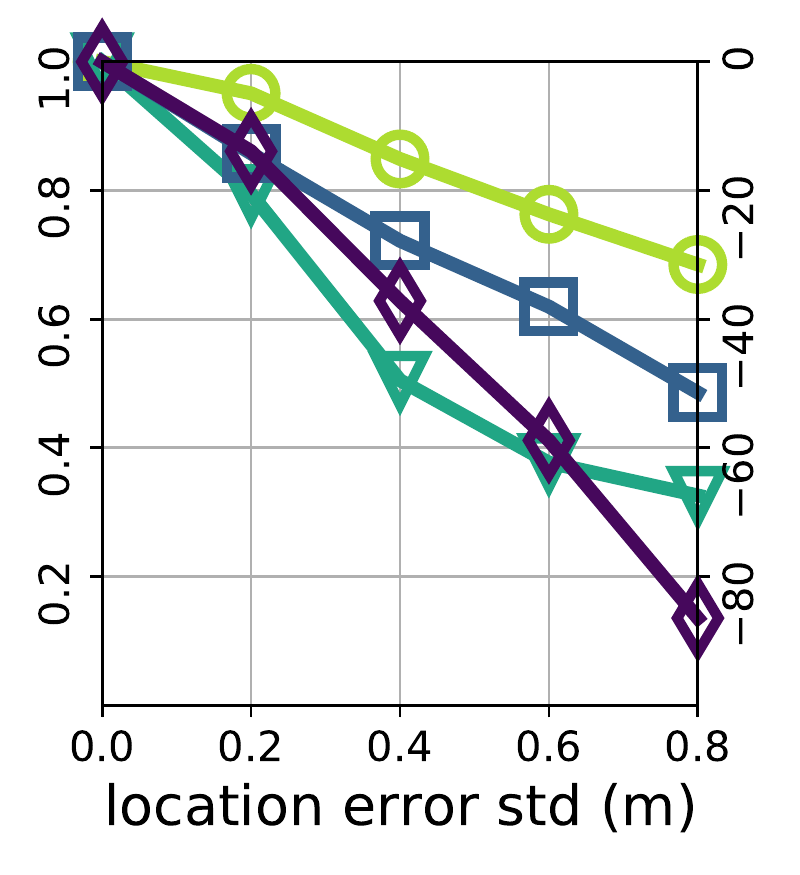}\\
   \vspace{-1mm}
        \includegraphics[width=\ew]{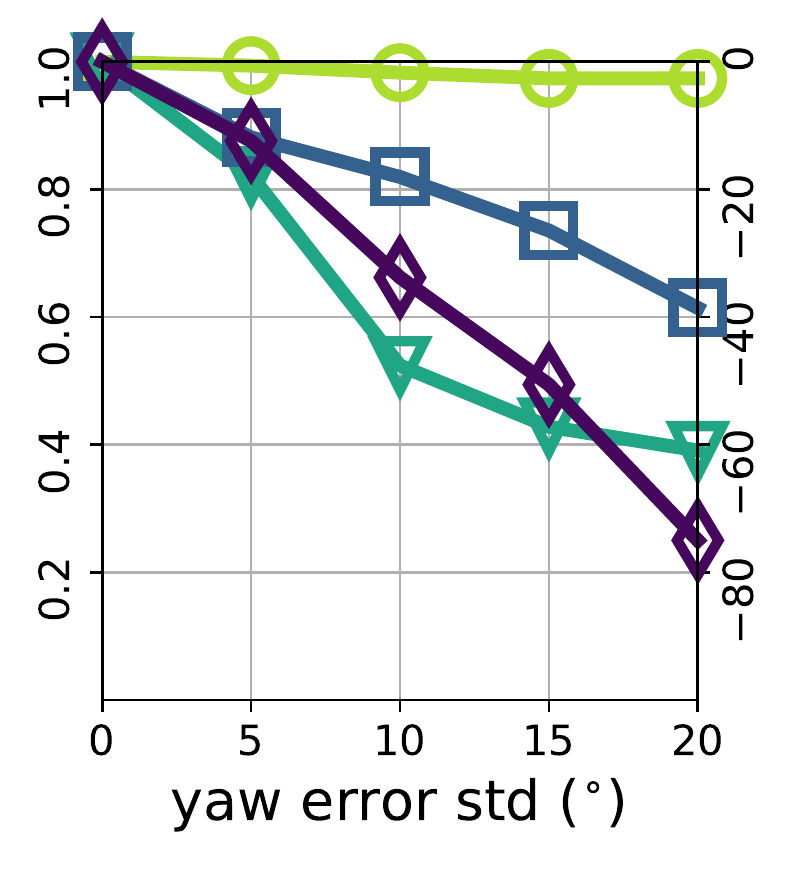}%
        \includegraphics[width=\ew]{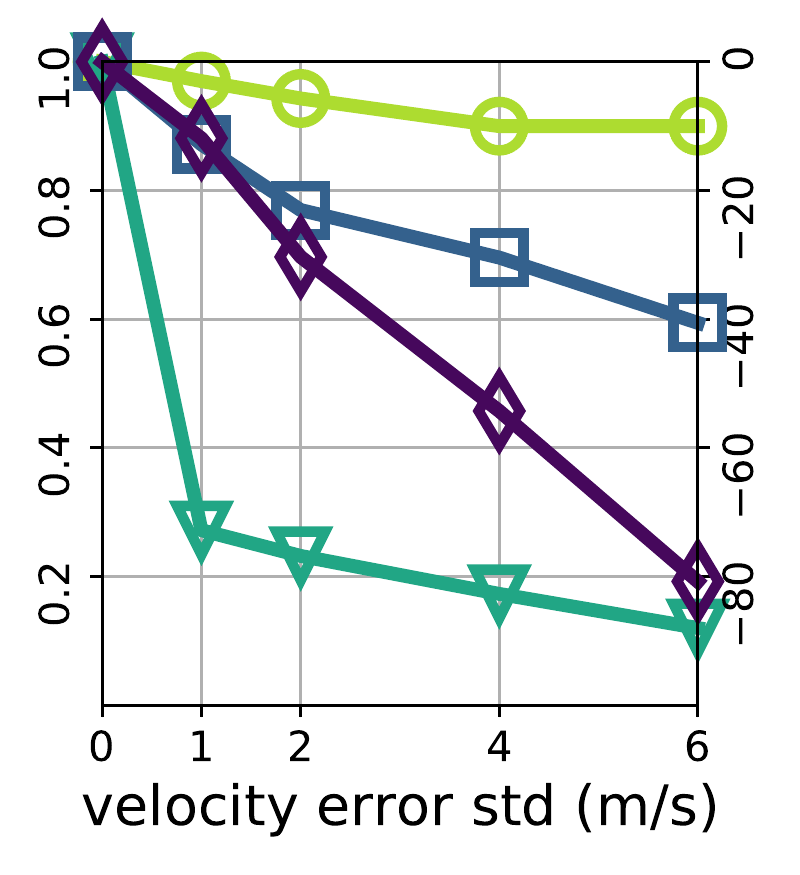}%
        \includegraphics[width=\ew]{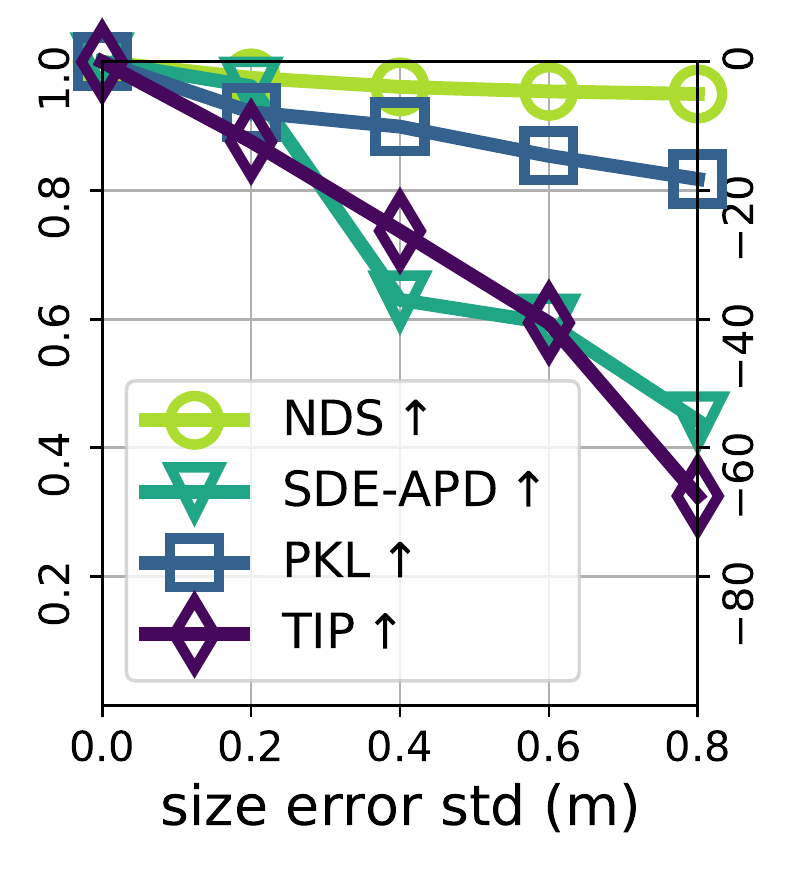}
   \vspace{-6mm}
	\figcaption{\label{fig.synthetic.noise}\figcapmaker{Comparison of metrics on different cases of synthetic noise.}
        The left~(right) vertical axes are for NDS and SDE-APD~(PKL and
        TIP)\footnotemark.
        }
\vspace{-2mm}
\end{figure}
\footnotetext{Note numerical results of different metrics (\eg~PKL and TIP) are {\bf not} directly
comparable \interse, although they may be plotted in the same scale for brevity.
Instead, all relevant
observations or conclusions are made from corresponding trends \perse.}
%%%%%%%%%%%%%%%%%%%%%%%%%%%%%%%%

%%%%%%%%%%%%%%%%%%%%%%%%%%%%%%%%
\def\ew{3.3in}
\begin{figure}[t]
   \centering
  \includegraphics[width=\ew]{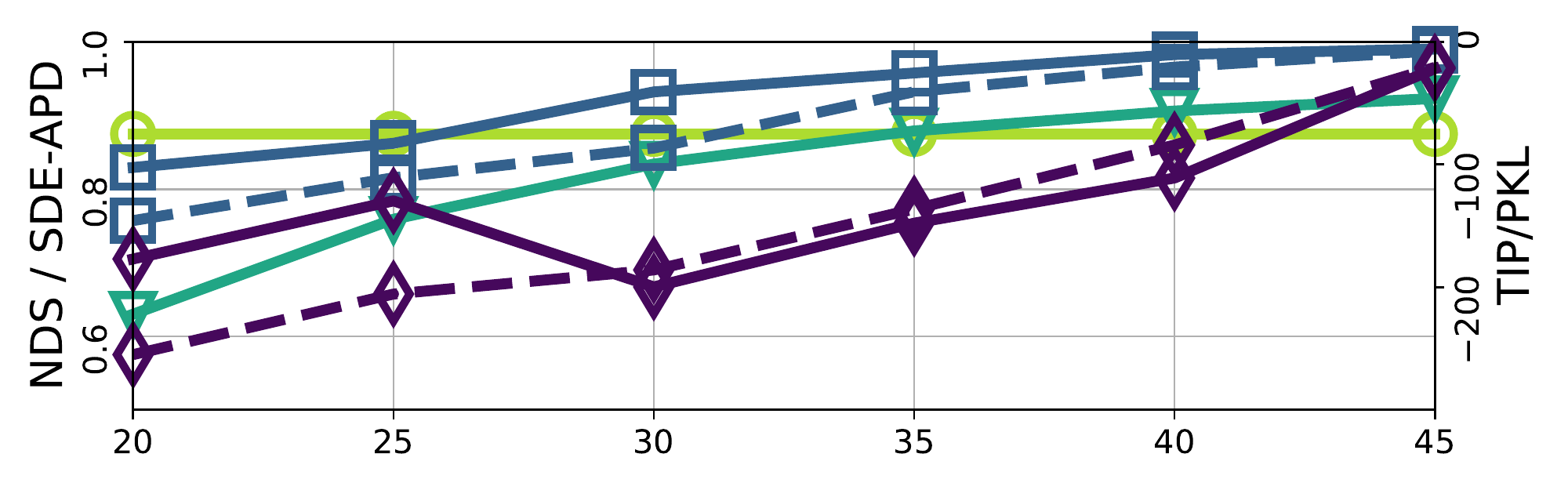}\\
   \vspace{-2mm}
   \includegraphics[width=\ew]{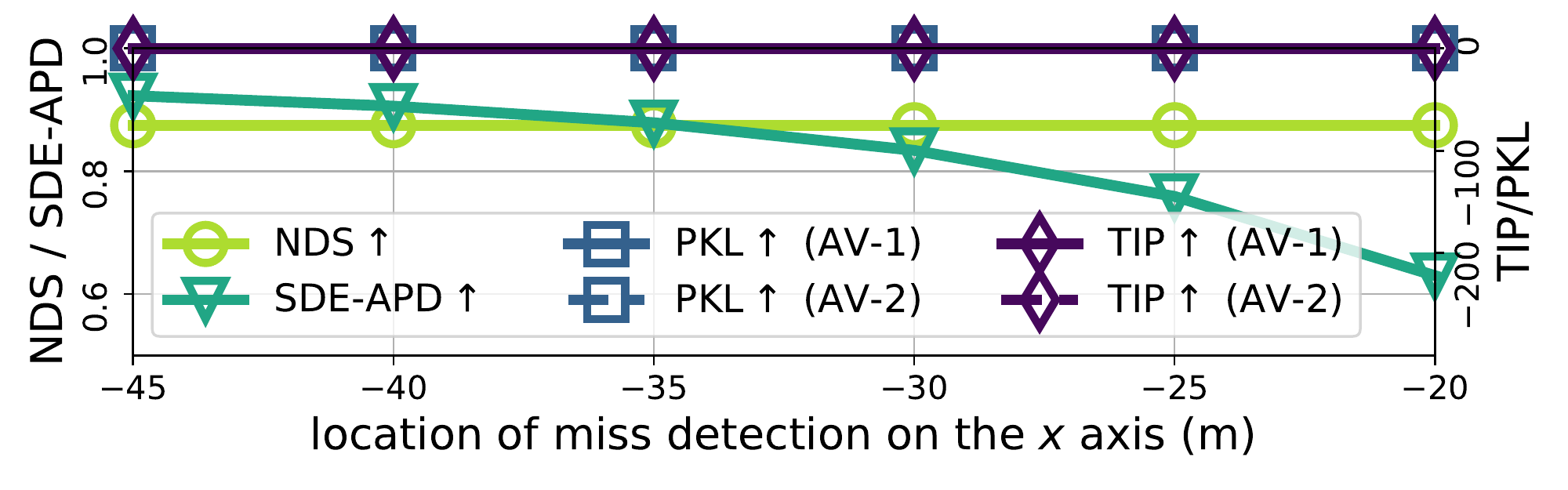}
   \vspace{-7mm}
   \figcaption{\label{fig.synthetic.md_demo}\figcapmaker{Metrics for AVs of different driving styles.}
     On the $x$-axis are 
     (i) a miss detected stationary obstacle;
     (ii) another stationary vehicle at $x$=50;
     and (iii) an AV at $x$=0 moving along the +$x$ direction at 14m/s. AV-1 (`jerk-averse') is
    optimised for driving comfort with braking capped at
    -4m/s\textsuperscript{2}, while AV-2 (`collision-averse') is for safety,
    which can brake as much as -6m/s\textsuperscript{2}. 
    The stopping distance is around 30m and 20m, respectively.
   }
\end{figure}
%%%%%%%%%%%%%%%%%%%%%%%%%%%%%%%%

%%%%%%%%%%%%%%%%%%%%%%%%%%%%%%%%
\setlength{\floatsep}{3mm}
\begin{figure*}[t]
      \centering
      \begin{minipage}[c]{0.33\textwidth}
        \hspace{-8mm}
        \includegraphics[width=\textwidth]{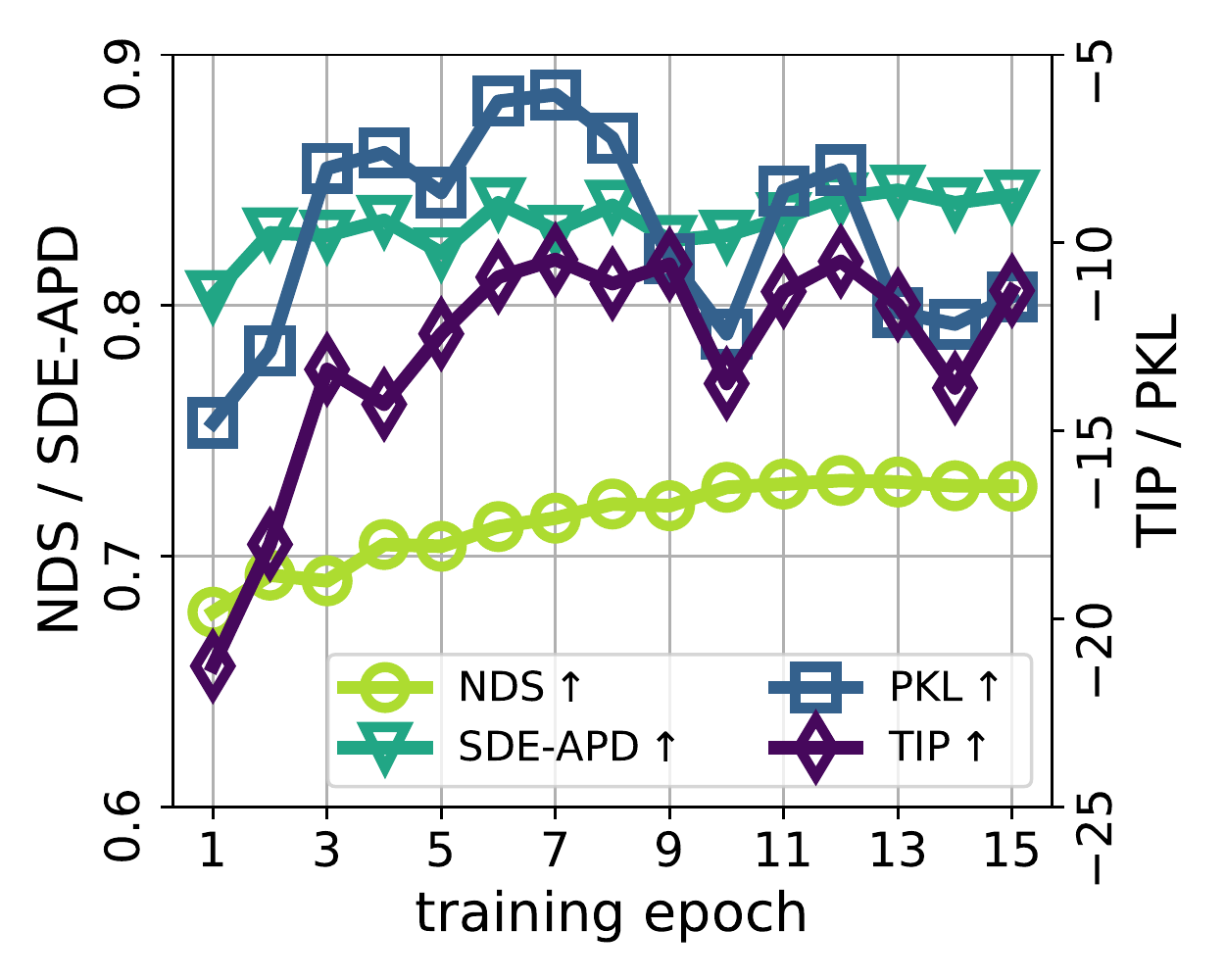}
      \end{minipage}
      \begin{minipage}[c]{0.27\textwidth}
        \hspace{-6mm}
        \includegraphics[width=\textwidth]{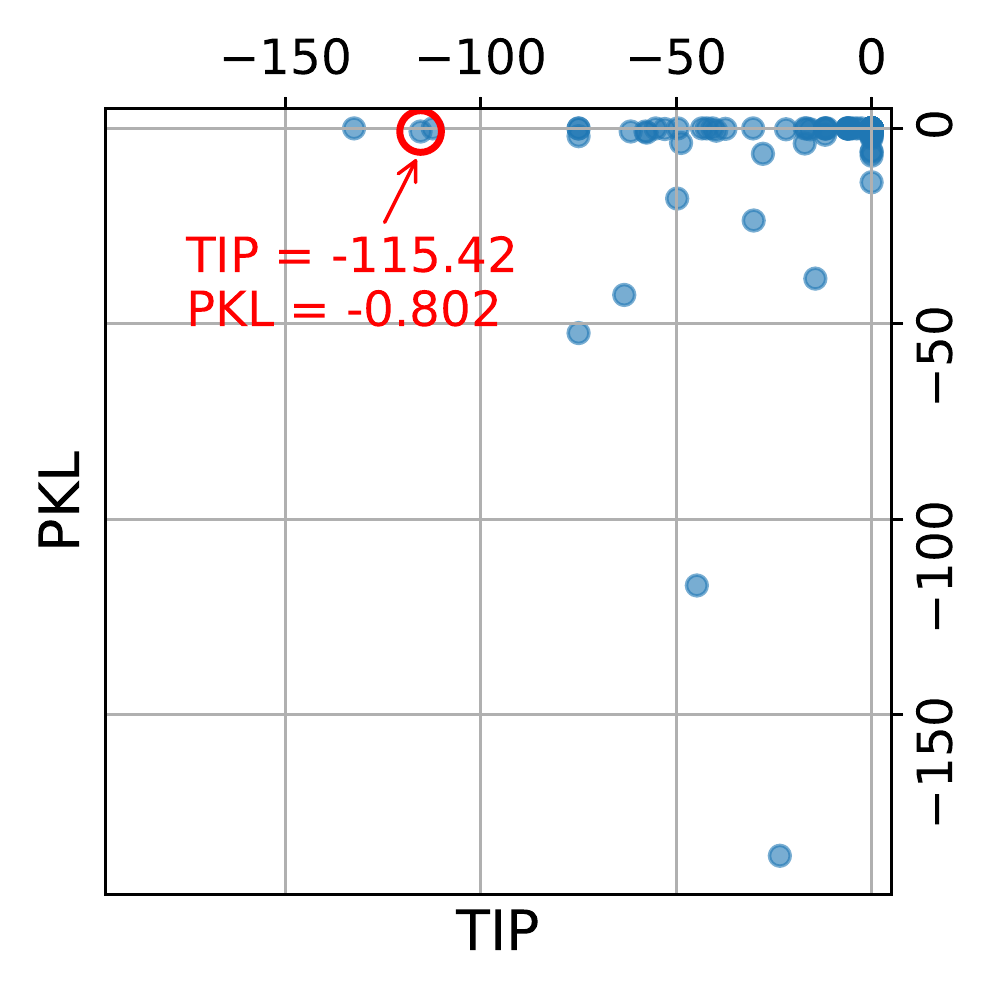}
      \end{minipage}
      \begin{minipage}[c]{0.32\textwidth}
        \includegraphics[width=1.1\textwidth]{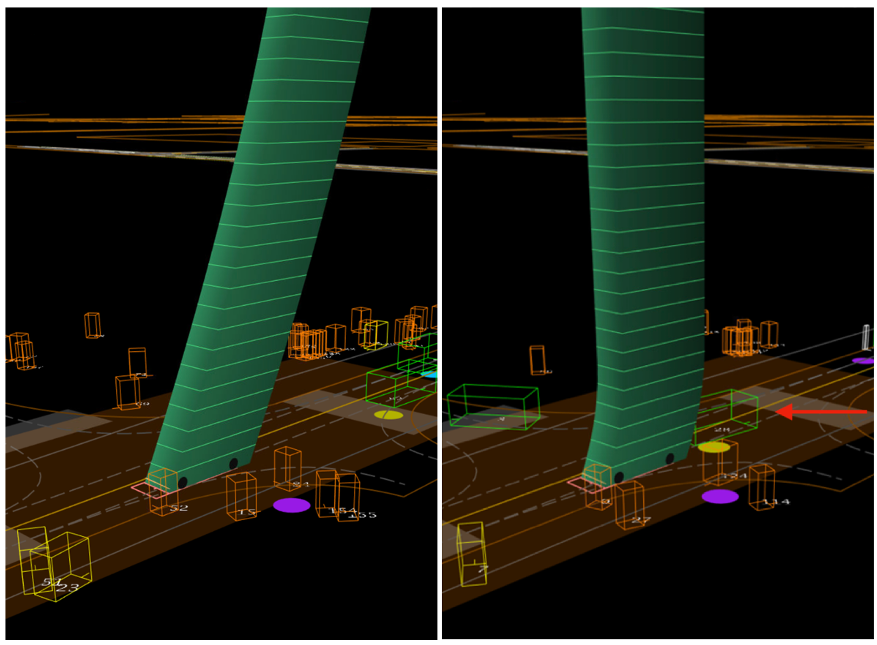}
      \end{minipage}
       \vspace{-2mm}
       \figcaption{\figcapmaker{Comparison of metrics on real data.}
        Left: metrics on different checkpoints during training.
        Middle: scatter plot of impacts of  perception noise measured by TIP and
        PKL. Note the number of data points close to the $x$-axis (PKL = 0),
        which correspond to critical errors in planning due to the perception
        noise captured by TIP yet missed by PKL since the AV behaviours are
        similar with and without the noise.
        Right~(best viewed in colour): the first one is the ground truth with the corresponding AV
        behaviour~(the spatiotemporal trajectory illustrated by the green
        tube with the $z$-axis as the temporal dimension); the second one shows 
        an outrageous error of a false positive (pointed to by the red arrow),
        which causes a jerk of $-76.4$m/s\textsuperscript{3} while the
        typical limit is around $-1.0$m/s\textsuperscript{3}~\citep{Wang2018},
        despite a mild change in behaviour per PKL. PKL and TIP of this case are
        highlighted by the red circle in the middle scatter plot.
        See details in
        \refapp{app:comparison.pkl}.
        }
\label{fig.real}
\vspace{-2mm}
\end{figure*}
%%%%%%%%%%%%%%%%%%%%%%%%%%%%%%%%

\vspace{-2mm}
\subsubsection{Reaction to Different Types of Noise}
\label{sec.exp.sync.noise}
\vspace{-1mm}
In total, six types of errors are considered. The false positives are tested by
adding `ghost' vehicles scattered within a 70m-by-30m box centred at the AV,
with motion properties randomly perturbed from it. The miss detection is
created by removing objects from the ground truth randomly with a certain
probability (\ie~miss detection rate). Other noises involving location, yaw,
velocity and size are sampled from zero-mean Gaussian with
different variances and added to corresponding properties of ground
truth. The results are shown in Figure~\ref{fig.synthetic.noise}.
While all metrics negatively correlate with all six types of
noises, NDS saturates in some cases~(\eg~velocity) due to its design.
SDE-APD, also handcrafted with parameters calibrated from particular data sources,
exhibits varying sensitivity at different
noise levels, especially for the velocity (computed by SDE-APD@$t$=1s),
as the default matching threshold 0.2m is easily overwhelmed by speed noise larger than 1m/s.
In comparison, planner-centric metrics like PKL and TIP, with little
manual engineering involved, render
more consistent sensitivity across the whole dynamic range of different noises.

\vspace{-2mm}
\subsubsection{Case Study with Different Planners}
\label{sec.exp.sync.case}
\vspace{-1mm}
We further investigate the behaviour of TIP with different planner settings. In a typical
miss detection scenario, we remove a stationary vehicle in front of or behind an AV moving forward,
as shown in~\reffigure{fig.synthetic.md_demo}.
Both SDE-APD and PKL consider closer miss detections, under any circumstances,
worse than further ones.
TIP, however, predicts that AV-1 regards the one at 30m as the worst: the collision is
    inevitable even if the obstacles at 20m and 25m are detected;
     yet the miss detection at 30m leads to {\it a collision that could have been (barely) avoided
  otherwise}.
  In contrast, no other metrics provide insights at this level of subtlety.
  This demonstrates the superior resolution of TIP in identifying critical events from the planning
perspective that would have been missed by all other baselines (especially
SDE-APD, which explicitly incorporates the belief that the closer the miss detection is,
the worse it is \perse).
  On the other hand, when the miss detection happens behind the AV, both TIP and PKL ignore
  its impact. NDS and SDE-APD, however, fail to distinguish errors on both sides of the AV, due to
their spatial or directional homogeneity by design (note the symmetry of them in both
  directions of the $x$-axis).
  
\vspace{-2mm}
\subsection{{Results on Real Data}}
\vspace{-1mm}
In the second set of experiments, we study the results from the real perception
module deployed on our AVs, which is exemplified by a 3D object detection network
that predicts the class, location, heading, velocity and size of objects from LiDAR
point clouds. TIP is independent of the specific detector and can be applied to
various methods
\citep{point-pillars,pv-rcnn,center-point,pillar-next}. 
We develop an effective and efficient pillar-based network, 
which is trained on 780K
LiDAR sweeps using annotations of vehicle, pedestrian and cyclist with a detection
range of [-67.2m, 124.8m]$\times$[-51.2m, 51.2m].

\vspace{-2mm}
\subsubsection{Training Checkpoints}
\vspace{-1mm}
A typical challenge in developing a perception model is to determine how much
training is needed to reach a satisfactory level of performance. Conventional
solutions require a variety of heterogeneous metrics to measure different
aspects of an algorithm (\eg~mean average precision
for detection, mean squared errors for motion properties). Recently,
unified metrics like NDS~\citep{nuscenes} are also proposed by manual
engineering, which hardly confirm the driving quality
improvement of a perception model change. In most cases, conclusions can
only be made from large-scale real road tests, which are extremely costly~\citep{Wachenfeld2016, 2017sljung}.

We evaluate the performance of our 3D object detection model on the same
test scenarios (without any artificial noise) as in~\refsection{sec:exp.synthetic}
and compare the model output against the ground truth.
The model is trained for 15 epochs,
with results reported on the left of~\reffigure{fig.real}.
Unsurprisingly, NDS tends to increase as the training progresses and the
final checkpoint models usually achieve the best performance since NDS combines
the errors that are aligned with the loss functions optimised during training.
When evaluated with the AV involved, however, the observation changes.
SDE-APD implies that the training seems to
struggle with improving results on close-by objects as the losses are dominated
by a large number of far-away yet more challenging objects. From either behaviour
or planning perspectives, TIP and PKL both indicate that the last checkpoint
model is not among the best possible models during training. Instead, models
somewhere in the middle of the training can {provide better autonomous driving
performance.} Actually, neither TIP nor PKL is improved significantly beyond the
7th epoch, suggesting that early termination of training may be even more
beneficial to driving. {More importantly, we notice that TIP disagrees
with PKL on scenarios across models of top performance,
and there are quite some critical cases identified by TIP yet missed by PKL.
The difference is illustrated in the middle of~\reffigure{fig.real} by the
scatter plot of randomly sampled scenarios, where PKL is almost zero
while TIP scores are nontrivial in many scenarios, suggesting the
drastic impacts of perception errors on the planning process despite
similar AV behaviours with or without these errors (see the example
on the right of~\reffigure{fig.real} and others discussed in~\refapp{app:comparison.pkl}).}

%%%%%%%%%%%%%%%%%%%%%%%%%%%%%%%%
\def\ew{1.05in}
\begin{figure*}[t]
\setlength\parindent{-3.5mm}
	\centering
	\includegraphics[width=0.2\linewidth]{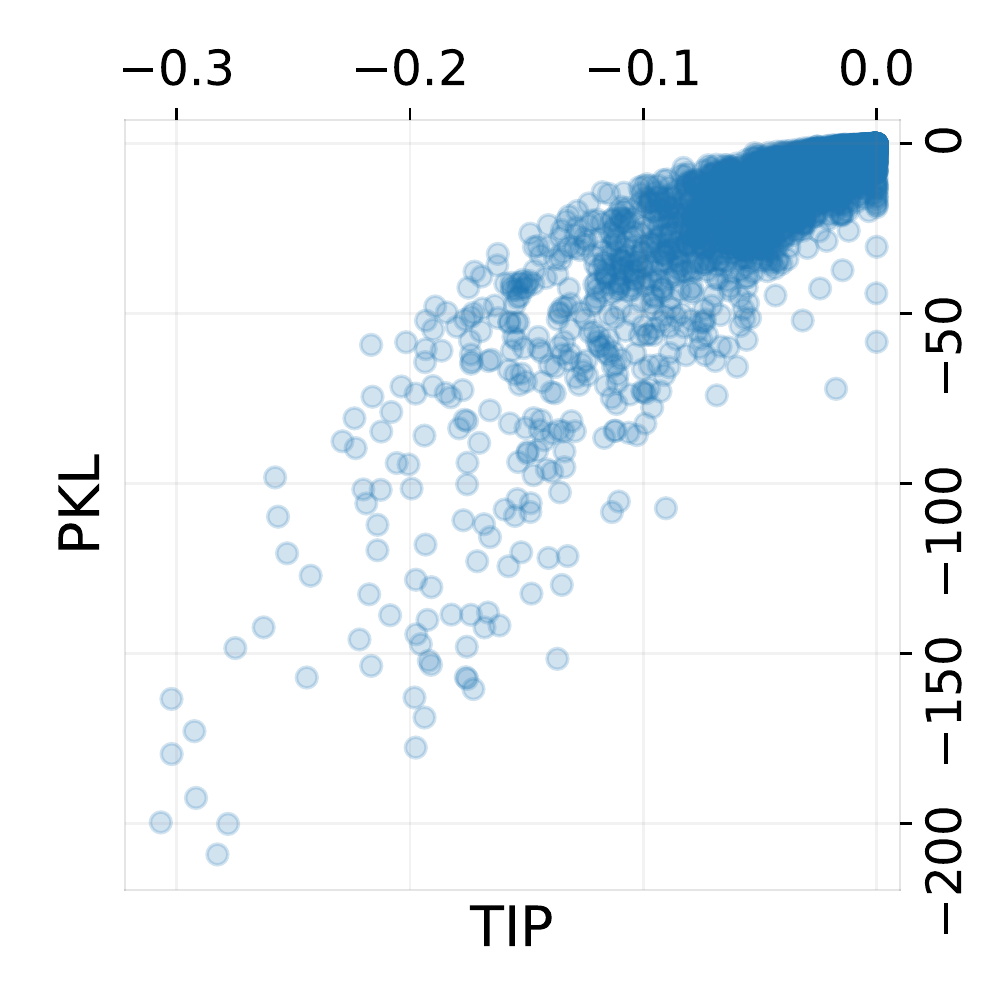} 
	\includegraphics[width=0.79\linewidth]{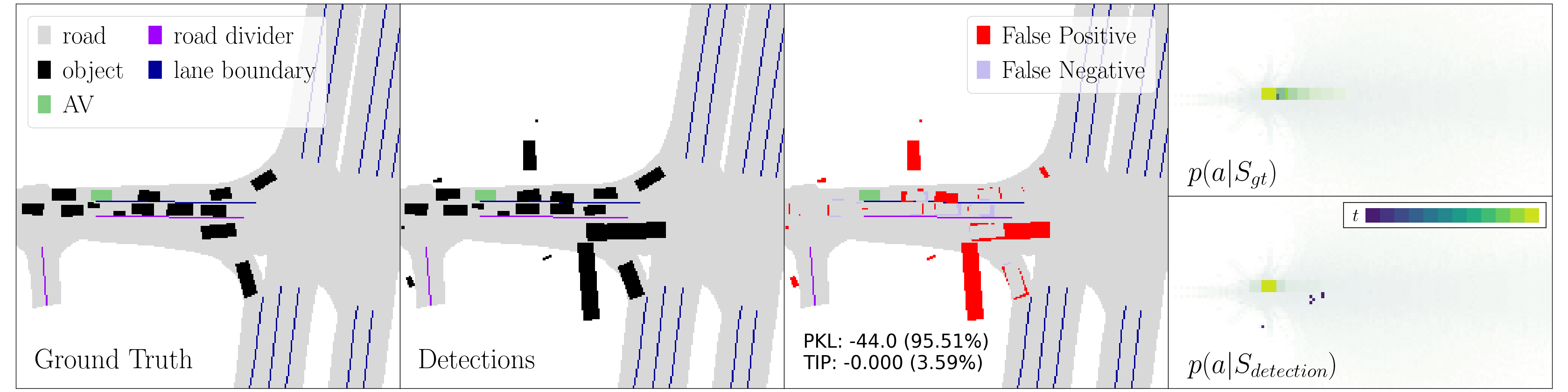}
   \vspace{-1mm}
	\figcaption{\label{fig.neural}\figcapmaker{Results of CBGS detector on nuScenes validation set.}
	Left: scatter plot of PKL and TIP scores (note that ranges of both metrics are different from previous experiments since the planners used are distinct).
	Middle and Right: a typical scenario where PKL deems a large impact on the planner from the perception noise
	 yet TIP considers it insignificant (score percentiles in the whole dataset are also shown in parentheses). Ground truth, detector outputs, their difference, and planner outputs (AV location distributions over time with one solid colour indicating the most likely locations at one time step, which are not plotted to the numeric scale for visual saliency enhancement) are shown, respectively.
     	}
\vspace{-2.5mm}
\end{figure*}
%%%%%%%%%%%%%%%%%%%%%%%%%%%%%%%%

\vspace{-2mm}
\subsubsection{More Perception Models}
\vspace{-1mm}
 \begin{table}[t]
\vspace{-1mm}
\tabcaption{
\label{tab:detectors} Comparison of different perception models.
}
\vspace{-3.5mm}
\begin{center}
\begin{small}
\begin{sc}
\begin{tabular}{c c c c c }
\bottomrule
  Detector 			& NDS$\uparrow$ 	& SDE-APD$\uparrow$ 	& PKL$\uparrow$ 	& TIP$\uparrow$ \\
\hline
  Pillar 	&0.730  		&0.843 		& -8.1 	& -10.5 	\\
  PillarNeXt-1F		& 0.693 		& 0.852		& -9.2	& -11.7 	\\
  PillarNeXt-5F	& 0.744 		& 0.878 		& -7.9 	& -9.1 	\\
\bottomrule
\end{tabular}
\end{sc}
\end{small}
\end{center}
 \end{table}

\begin{table}[t]
  \vspace{-1mm}
  \tabcaption{\label{tab:subjective} Metric favour rate$\uparrow$ by the subjective evaluation.}
    \vspace{-3.5mm}
  \begin{center}
  \begin{small}
  \begin{sc}
  \begin{tabular}{c | c c c}
  \bottomrule
    Metric 		&  NDS & SDE-APD & PKL \\
  \hline
  TIP$^\dagger$ 		& 82\%$^\dagger$\vs 18\% & 66\%$^\dagger$\vs 34\% & 61\%$^\dagger$\vs 39\%\\
  \bottomrule
  \end{tabular}
  \end{sc}
  \end{small}
  \end{center}
  \vskip -0.11in
  \end{table}

To evaluate other 3D detectors,
we implement two more models with the recent
PillarNeXt~\citep{pillar-next} 
as the basic
detector. The first one (PillarNeXt-1F) uses the point cloud only from the
current frame for prediction, while the second one (PillarNeXt-5F) leverages 5
consecutive frames around the current one. Results are reported
in~\reftable{tab:detectors}. Both models
have better performance by SDE-APD. PillarNeXt-1F, however, fails to
produce precise velocity from single-frame observation (not reflected by SDE-APD), leading to
an inferior performance by the other three metrics.
PillarNeXt-5F delivers overall best results across all
metrics, despite marginal gaps by PKL and NDS.

\vspace{-1mm}
\subsubsection{Subjective Evaluation}
\vspace{-1mm}
To further justify the soundness of the proposed approach on the scenario level,
we also implement a set of subjective evaluations similar to that in
\citep{Philion2020}. We collect 258 scenario pairs with actual
perception noises and check whether TIP, PKL, SDE-APD or NDS disagree on the relative
severity, \ie~one believes the perception error in scenario A is worse than
that in scenario B while the other one thinks alternatively. These pairs of
scenarios are compared by 10 human drivers to decide
which is worse subjectively. The result reported
in~\reftable{tab:subjective} suggests that human drivers side with TIP more over
the other three baselines.

\vspace{-2mm}
\subsection{Application to Neural Planners}
\label{sec:neural}
\vspace{-1mm}
The proposed framework is also applicable to neural planners with implicitly 
derived behaviour cost or likelihood functions for inference such as \citep{Bansal2019, Zeng2019, Philion2020}.
For this, we implement TIP scoring on the 
neural planner from~\citep{Philion2020}, where the output trajectory  
\pdf~$\probb(\action|\staterv)$ is adopted in lieu of the utility function $\Utility(\staterv, \action)$ for TIP. 
Given any perception input, the planner produces a distribution of AV future actions 
 $\probb(\action|\staterv)$
and the one with the highest  probability (density) is chosen as the AV behaviour
$\action^*=\argmax_\action\probb(\action|\staterv)$.

Under this setting, PKL and TIP evaluate the impact of perception noise on the planner
 with different nuances, as reflected by the results shown in \reffigure{fig.neural}, 
 where the PKL and TIP scores for the CBGS 
 detector~\citep{zhu2019class}
on the validation set of nuScenes 3D object detection task~\citep{nuscenes} 
 are presented.  
The former
considers feasible actions of {all road vehicles} and aggregates the difference 
between the action distribution given ground truth and perception inputs 
across the whole action space.
The latter, in contrast, focuses on the optimal action $\action^*$ {\it the AV actually 
executes} (given the ground truth input and subject to kinetic and kinematic constraints) 
and evaluates the reduction in AV's favourability 
on $\action^*$ over any other candidate actions given the two different inputs.        
Consequently, TIP captures a nontrivial number of scenarios where perception noises 
affect the behaviour of some general road vehicles but impact not necessarily 
that much on the AV \perse.  
See \refapp{app:neuralplanner} for more discussion.

\vspace{-2mm}
\section{Notes on Dependence on the Planner}
\vspace{-1mm}
The proposed framework relies on a planner's reactions to input noises to evaluate 
perception, a prominent property for all planning-centric 
metrics~\citep{Philion2020, Ivanovic2021}.    
Due to this dependence on the planner, evaluation results of these metrics on the 
same perception input may change as the underlying planner varies.  
While not necessarily a drawback, it does incur some extra cost to ensure 
proper application of these metrics. 
Most importantly, the planner should be sufficiently verified before being deployed with the 
metrics for evaluation, either by validation on 
benchmarks~\citep{Philion2020}, 
virtual simulation~\citep{Dosovitskiy2017}, 
or real world road test as for our planner in \refsection{sec:settings}.
In addition, all interpretations of the result should be made in the 
specific context of the planner employed, and any observations are 
planner-bound, \eg~numeric scores from the same metric should only be compared 
against those from the same planner \perse.  

\vspace{-2mm}
\section{Conclusion}
\vspace{-1mm}
In this work, we have proposed TIP, a principled framework to evaluate perception from the
planning perspective for autonomous driving. TIP explicitly exploits
 properties of utility-based planners and effectively identifies perception
noises that may cause large planning changes in the context of expected utility
maximisation. Extensive experiments on both synthetic and real data confirm that
TIP is capable of distinguishing perception errors that would not be
identified by the conventional and ego-centric metrics, or those exclusively
focusing on behaviours output from the planner.

{
\bibliography{ref}
\bibliographystyle{icml2023mod}
}

% APPENDIX
\onecolumn
\newpage
\appendix
\addcontentsline{toc}{section}{Appendix} 
\part{Appendix} 

\section{Planning-Critical Errors}
\label{sec:pce}

\begin{figure*}[t]
     \centering
     \begin{subfigure}[b]{0.39\textwidth}
         \centering
         \includegraphics[width=0.95\textwidth]{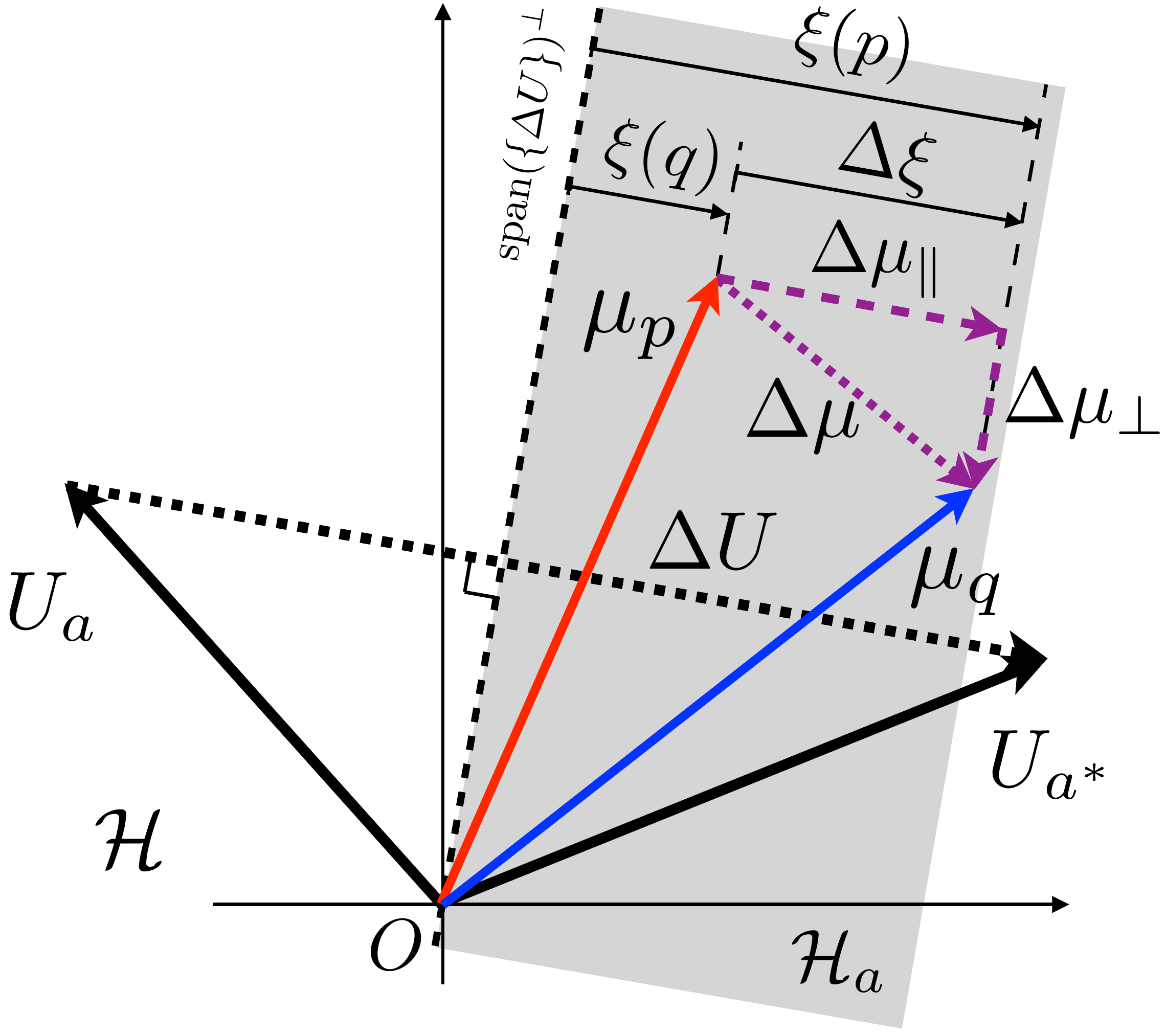}
         \caption{Illustration of EUM in $\Hspace$ for PCE ($\unitvec_{\Delta\mu_\parallel}=\unitvec_{\Delta\Utility}$).}
         \label{fig.H.pbe}
     \end{subfigure}
     \hfill
     \begin{subfigure}[b]{0.6\textwidth}
         \centering
         \includegraphics[width=1.0\textwidth]{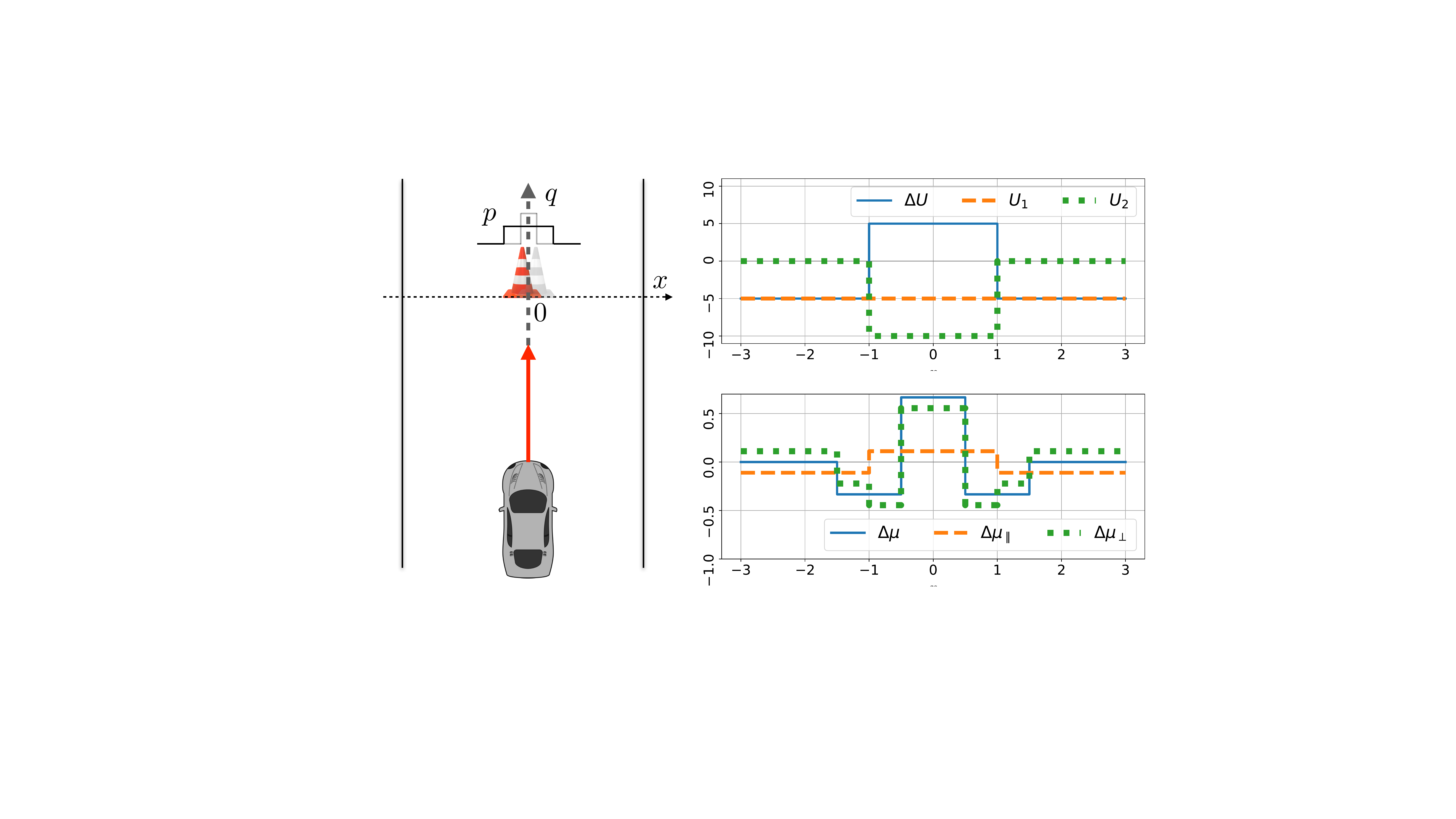}
         \caption{An example of PIE $\Delta\mu_\perp$ and PCE $\Delta\mu_\parallel$ when ${\footnotesize\langle{\Delta\mu_\parallel},{\Delta\Utility}\rangle}>0$ .}
         \label{fig.pbe.example}
     \end{subfigure}
        \caption{
        \figcapmaker{Illustration of TIP for the planning-critical error~(PCE).}
        		In ($a$), $\Delta\Utility=\Utility_{\action^*}-\Utility_{\action}$ defines the
		behaviour direction;
		$\xi$ represents the preference score; $\mu_\probb$ and
		$\mu_\qrobb$ are the embeddings of the ground truth and perception result, respectively;
		$\Delta\mu$ is the perception error, which is
		decomposed into the planning-critical error~(PCE) $\Delta\mu_\parallel$, and
		the planning-invariant error~(PIE) $\Delta\mu_\perp$; and the shaded area
		corresponds to $\Hspace_\action$.
		Note that ${\footnotesize\langle{\Delta\mu_\parallel},{\Delta\Utility}\rangle}>0$ in this case.
		In ($b$), an AV is moving forward on a road of width $6$m,
		A cone is in front of the AV, with its position distributed
		on a line across the road (the $x$ axis). The ground truth distribution
		$\probb$ is $\uniform_{[-1.5,1.5]}$, a uniform distribution with support
		$[-1.5,1.5]$, while the perception believes its location (distributed as
		$\qrobb$) is $\uniform_{[-0.5,0.5]}$. The $2$m-wide AV has two action options:
		(\romannumeral 1) to make a hard brake and come to a full stop before the $x$-axis ($\action^*$,
		the red solid line with an arrowhead), and the utility function is a constant
		$\Utility_1(x)=-5$ with $x$ being the position of the
		cone (loss of hard braking is identical regardless of the cone
		position);
		(\romannumeral 2) to move forward
		($\action$, the grey dashed line with an arrowhead), and the utility function is
		$\Utility_2(x) = -10\cdot\1{x\in[-1, 1]}$ (only large loss for collision with the cone).
		The $\Delta\Utility$ and $\Delta\mu$ are
		illustrated in the top right, while the decomposition of PIE
		$\Delta\mu_\perp$ and PCE $\Delta\mu_\parallel$ are in the bottom right.
		Note that, $\Delta\mu_\parallel$ is of the same shape as $\Delta\Utility$~(thus
		${\footnotesize\langle{\Delta\mu_\parallel},{\Delta\Utility}\rangle}>0$), and
		$\innerprod{\Delta\Utility}{\Delta\mu_\perp} = 0$.
        }
        \label{fig.pbe}
\end{figure*}

When $\unitvec_{\Delta\mu_\parallel}=\unitvec_{\Delta\Utility}$,
the change in preference score $\Delta\xi(\action^*, \action; \qrobb, \probb)$
of~\refeqn{eq:decomp} is positive, suggesting that the difference between the expected reward by executing $
\action^*$ and that of $\action$ is even larger with noisy perception input $\qrobb$ than with the ground
truth $\probb$, \ie~the planner is even more confident in choosing $\action^*$ over $\action$ given the erroneous $\qrobb$.
The breakdown of EUM and an example is shown in~\reffigure{fig.pbe} for this case.
Notably, in the example, although there is a non-trivial probability (1/3) that the AV may
 not collide with the
cone in the ground truth (when the cone is in the range $[-1.5,-1.0]\cup[1.0,1.5]$) even if it moves forward
(action $\action$), the risk is still too high given the cost of collision, and hard braking (action $\action^*$) is
preferred for peace of mind since $\xi(\probb; \action^*, \action)=\frac{5}{3}$. The noisy perception, on the other
hand, predicts that the AV will
almost surely collide with the cone if it moves forward, which makes the AV $100\%$ sure that coming to a
stop is absolutely necessary since $\xi(\qrobb; \action^*, \action)=5$.
Note that, insights into the nature of this type of perception error provided by
our proposed analytical framework are not possible from other baselines like NDS, SDE-APD, and PKL, which
 assign non-positive impacts to all kinds of errors.

\section{Scenario Collection}
\label{app:data}
The scenarios used in this work are curated from AV road tests in real world from
public roads in urban areas of megacities, \eg~central business districts, populated residential communities, major commercial areas, etc. 
Each scenario is a 10s-long excerpt extracted from a continuous interval
of a road test, which consists of
(i) all raw data recordings (LiDAR point clouds, camera images, positioning signals, \etc) from 
the road test within the interval,
and (ii) the portion of offline generated high-definition (HD) and birds-eye view (BEV) raster maps that cover the field of perception during the interval.
The duration of a road test ranges from tens of minutes to several hours,
and covers various times on both weekdays and weekends from early morning
till late night during a period of more than one year, providing a rich
blending in
weather condition (\eg~sunny, cloudy, rainy, and snowy),
traffic intensity (\eg~congested highways during rush hours and crowded streets on holidays),
road participant diversity (\eg~private cars, cyclists, pedestrians, and emergency vehicles), and so forth.
The scenarios are selected from non-trivial situations (\ie~those
with few traffic participants are filtered out)
with a balance in AV motion speed, diversity of traffic participants, weather,
geographical locations, \etc

\section{Autonomous Vehicle Planner}
\label{sec:planner}

%%%%%%%%%%%%%%%%%%%%%%%%%%%%%%
\begin{figure*}[t]
		\centering
		\includegraphics[width=\linewidth]{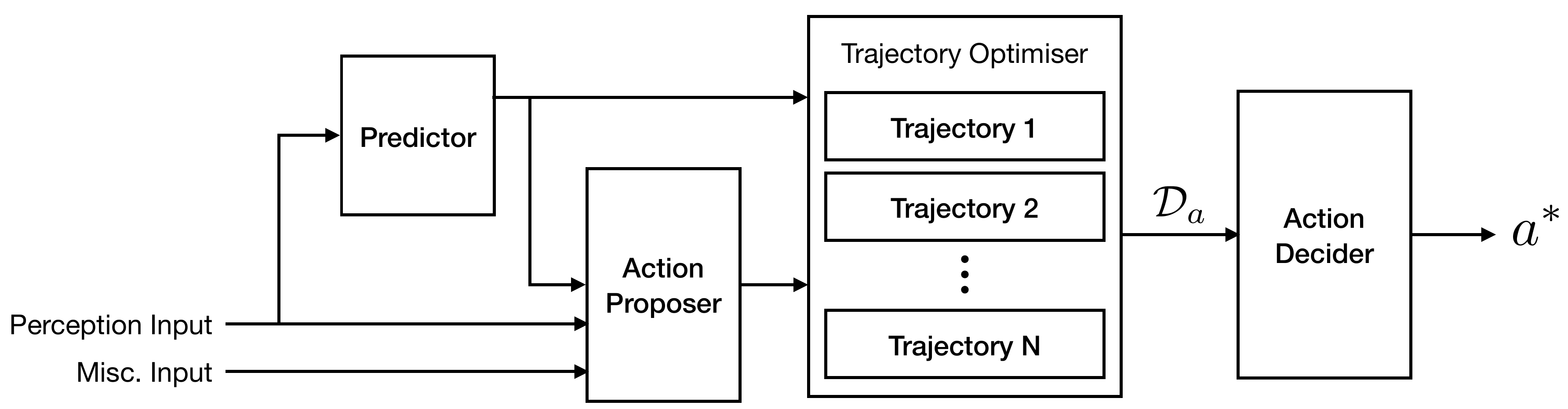}
		\caption{
		\figcapmaker{Diagram of the major components in the planner used in the experiment.}
		}
		\label{fig:planner}
\end{figure*}
%%%%%%%%%%%%%%%%%%%%%%%%%%%%%%

\begin{table*}[t]
\caption{Comparison of perception metrics for autonomous driving.}
\label{tab:merics}
\hspace{0.03cm}
\begin{footnotesize}
\begin{tabular}{cccccc}
\toprule
{\sc Metric}	& \tabincell{c}{NDS\\\citep{nuscenes}}	& \tabincell{c}{SDE-APD\\\citep{deng2021revisiting}}	& \tabincell{c}{PKL\\\citep{Philion2020}}	& \tabincell{c}{IPA\\\citep{Ivanovic2021}}	& TIP \\
\toprule
\tabincell{c}{Metric\\
Parametri-\\
sation} 	& Manual	& \tabincell{c}{Manual\\+ calibration}	& {None}	& {None}	& {None}\\
\midrule
\tabincell{c}{Evaluable\\Error Types}	& \tabincell{c}{Detection,\\category,\\velocity,\\heading,\\localisation,\\size}	& \tabincell{c}{Vehicle size,\\heading,\\location}	&{\tabincell{c}{Any perception\\inputs to\\the planner}}	& \tabincell{c}{Any perception inputs\\to the planner in the\\differentiable\\cost terms}	&{\tabincell{c}{Any\\perception\\inputs to\\the planner}} \\
\midrule
\tabincell{c}{Perception\\Input\\Represen-\\tation}	& Deterministic	& Deterministic	& Deterministic	& Deterministic	&{\tabincell{c}{Either\\deterministic\\or\\probabilistic}}\\
\midrule
\tabincell{c}{Reflecting\\Actual\\Severity?}	&  \tabincell{c}{Not\\necessary}	& \tabincell{c}{Not\\necessary}	& Partially	 & \tabincell{c}{For small local\\errors in the\\differentiable\\terms only} & {Yes}\\
\midrule
\tabincell{c}{Planner\\Dependence}	&None	&None	&\tabincell{c}{Planners with\\probabilistic \\trajectory\\output}	&\tabincell{c}{Planners with\\differentiable cost\\functions}	&\tabincell{c}{Planners with \\action-state\\utility\\functions}\\
\midrule
\tabincell{c}{Planner\\Specifi-\\cation}	&None	&None	&\tabincell{c}{Training data,\\planner network,\\learning algorithm\\(for neural planners)}	&Cost function weights	&\tabincell{c}{Utility function\\parameters and\\weights}\\
\midrule
\tabincell{c}{Planner for\\Empirical\\Study}	& None	& None	&\tabincell{c}{End-to-end planning\\architectures trained\\on nuScenes training\\set}	&\tabincell{c}{Learnt from\\human driving records\\in nuScenes}	&\tabincell{c}{Tuned/learnt\\from real\\world road\\test scenarios\\and human\\driving records}\\
\midrule
\tabincell{c}{Planner\\Validation}	&None	&None	&\tabincell{c}{Validated on 4k\\nuScenes trajectories\\(<200 miles in\\total length)}	&\tabincell{c}{Validated on 4k\\nuScenes trajectories\\(<200 miles in\\total length)}	&{\tabincell{c}{Validated on\\100k miles real\\world urban\\road test\\(>100 MPI)}}\\
\bottomrule
\end{tabular}
\end{footnotesize}
\end{table*}

Our planner is designed to control SAE\footnote{SAE International, formerly named the Society of Automotive Engineers.} Level 4 AVs operating in urban areas of major modern cities.
Its modularised architecture 
consists of four major components as illustrated in~\reffigure{fig:planner}:
\begin{itemize}
\item The {\it predictor} infers the motion information $\state_{m}$ in the future (\ie~$t>0$) for all dynamic road objects from perception input history (\ie~$t\leqslant0$) up to the planning time (\ie~$t=0$).
\item The {\it action proposer} analyses the current environment at the planning time from (i) the perception input, (ii)
future object motion input, and (iii) other input signals (\eg~localisation, traffic lights, semantic maps, routing path, \etc),
and proposes various sets of behaviours (\eg~`go straight' and `lane change') for the AV with an initial feasible spatiotemporal trajectory for each set.
\item The {\it trajectory optimiser} takes the results of the above components as input and finds the optimal spatiotemporal trajectory for each behaviour set by numerically solving an optimisation problem with the initial feasible trajectory from the proposer as the starting point.
\item The optimal trajectories from all behaviour sets are then submitted to the {\it action decider}, which assembles all
information to evaluate the utilities of different candidate actions (with corresponding optimal spatiotemporal trajectories),  and makes the final decision on $\action^*$.
\end{itemize}

The utility function $\Utility(\action, \state)$ of the planner is of the general form
$$
\Utility(\state, \action) = \sum\nolimits_i\prod\nolimits_{j=1}^{n_j}\lambda_{ij} \Utility_{ij}(\state, \action) + \Utility_\state(\state) + \Utility_\action(\action),
$$
where $\{\lambda\}$ are the (static) coefficients, the atomic element function $\Utility_i$ depending on
both $\action$ and $\state$ characterises the “compatibility” of action $\action$ and scenario $\state$, $\Utility_\state(\state)$ depicts the current environment, and $\Utility_\action(\action)$ evaluates the quality of the action.
These terms can be categorised into the following groups.
\begin{itemize}
\item The {\it smooth motion} group encourages motion without abrupt change in acceleration and penalises large jerks (\ie~the derivative of acceleration).

\item The {\it safety distance to obstacles} group is designed to keep the AV away from other
road objects to minimise the collision likelihoods and guarantee leeway for control. This distance is defined as the $\Lnorm{2}$ distance between the AV spatiotemporal sweeping contour and a foreign object on the road.

\item The {\it legal motion satisfaction} group is designed to enforce the AV to strictly follow all applicable traffic rules when in motion.
For instance, the cost of crossing solid yellow lines is made significant such that the behaviour is prohibited unless a collision cannot be avoided otherwise.
Some other legal options also come at certain costs to discourage high-risk behaviours (\eg~lane changes in crowded scenes).

\item The {\it progress to the destination} group aims to guide the AV to achieve goals in distant horizons and reach the final destination.

\end{itemize}

The aforementioned planner deployed onboard our AVs has gone through rigorous road tests in urban areas of major cities
with millions of population.
Results from 10,000-mile weekly road tests indicate that the planner achieves
111.3 miles per intervention (MPI), confirming that the planner used in this work
is a reasonable and validated one.

\section{More Comparisons to Related Metrics}
\label{sec:more}

In comparison to the other baseline metrics, \eg~PKL~\citep{Philion2020} and IPA~\citep{Ivanovic2021} 
that are recently proposed for evaluating perception in the context of autonomous driving, our approach 
provides a universal and principled solution to evaluate the impact of perception noises from the perspective 
of the planning process of an AV.
Highlights of comparison across these two and other perception metrics for autonomous driving 
are summarised in~\reftable{tab:merics}. 

\subsection{Comparison with PKL}
\label{app:comparison.pkl}

\begin{figure*}[t]
     \centering
     \begin{subfigure}[b]{0.35\textwidth}
         \hspace{-6mm}
         \includegraphics[width=1.1\textwidth]{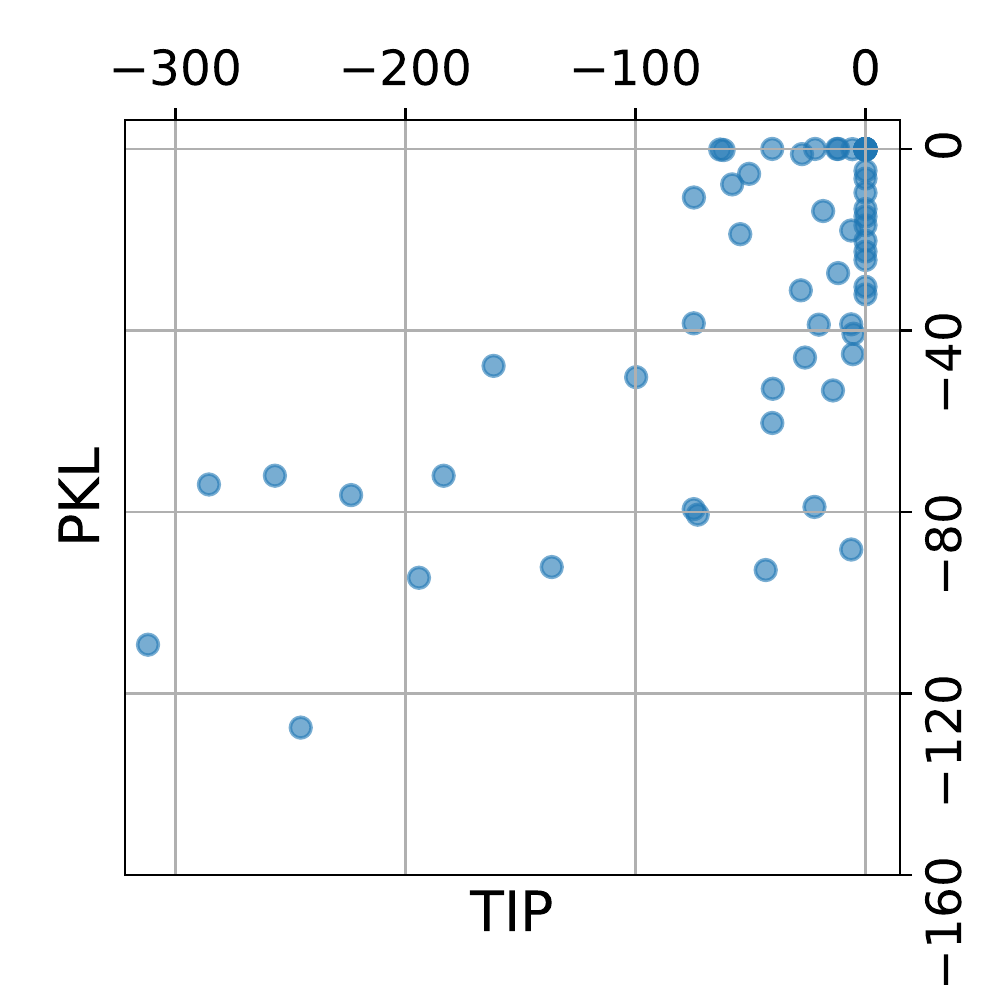}
         \caption{Scatter plot of PKL and TIP scores.}
         \label{fig:synthetic_fp.scatter}
     \end{subfigure}
     \hfill
     \begin{subfigure}[b]{0.64\textwidth}
         \centering
         \includegraphics[width=0.85\textwidth]{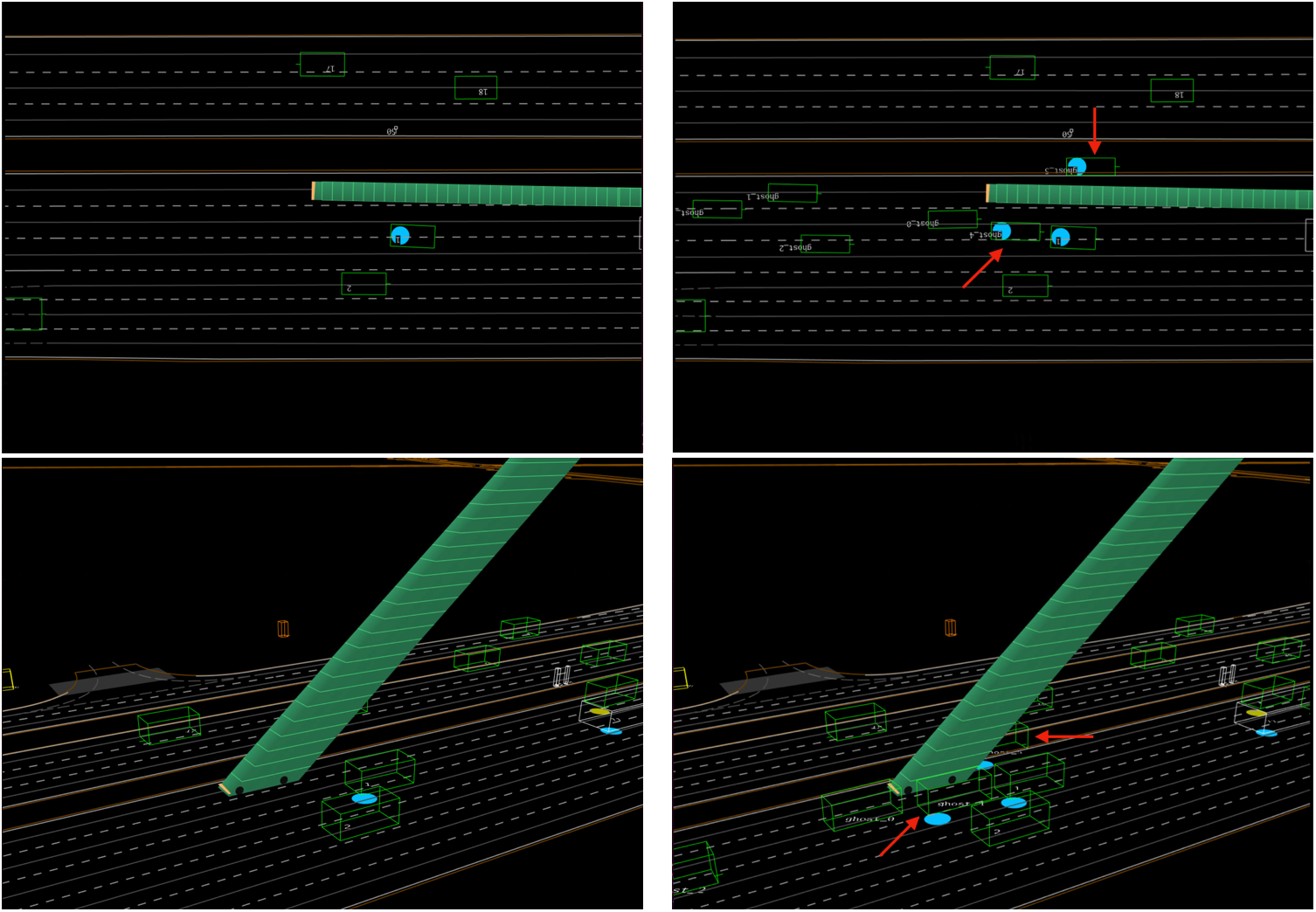}
         \caption{Illustration of AV behaviours in the ground truth and synthetic scenes with
		false positives.}
         \label{fig:synthetic_fp.example}
     \end{subfigure}
        \caption{
        \figcapmaker{Result on the false positive synthetic data~(best viewed in colour).}
        		In ($a$), the data points (downsampled for clarity) close to $x$-axis (PKL = 0) correspond to the cases where the AV behaviours under the ground truth and noisy perception inputs are identical. The data points close to $y$-axis (TIP = 0) correspond
		to the cases where the AV planning preference between the optimal action and others is identical under
		ground truth and noisy perception inputs.
		Note the number of cases where TIP disagrees with PKL on the impact on AV planning.
		In ($b$), The green tube represents the spatiotemporal trajectory of the AV with the z-axis as
		the temporal dimension (same for the rest).
		Bold solid lines are the boundary of driving areas (\eg~curbs, vegetarian zoom dividers),
		while light solid lines are the centre lines of vehicle lanes with dashed lines as the lane boundaries.
		Road objects are marked with 3D bounding boxes in green.
		Sub-figures in the first (second) row are birds-eye view (side view) of the scene and sub-figures
		in the left (right) column correspond to ground truth (noisy) perception input (same for the rest).
		In this case, the AV intends to move forward under the ground truth perception input (left column);
		in the presence of perception input noise (right column), the AV behaviour remains almost unchanged
		(PKL = -0.248), since two false positive vehicles (pointed by red arrows) on both sides force the AV
		to keep moving straight, yet the close-to-object cost (safety distance to road obstacles) has changed
		considerably during planning, leading to a TIP score of -61.654.
        }
        \label{fig:synthetic_fp}
\end{figure*}

More empirical results are provided to better understand the difference between the proposed TIP and PKL~\citep{Philion2020}.

 {\bf Results on Synthetic Data}.
\reffigure{fig:synthetic_fp.scatter} demonstrates a scatter plot for scene-wise TIP and PKL results on the synthetic data generated as described in Section 5.2 with 6 false positives per scene.
It is observed that some results are very close to either $x$- or $y$-axis, suggesting that TIP and PKL deviate in deciding if a perception error (\ie~false positive) is crucial to planning in these cases.
A typical scenario of such disagreement is shown in \reffigure{fig:synthetic_fp.example}, where
the behaviour of the AV does not change significantly with ground-truth or noisy perception inputs (PKL = -0.248),
yet the planning process has changed quite a lot (TIP = -61.654) due to the affinity of false positive objects that have drastically change the planning cost to close objects. In this case, TIP is capable of detecting serious perception errors that PKL fails to identify.

\begin{figure*}[t]
\def \eh{2.4in}
		\centering
		\includegraphics[height=\eh]{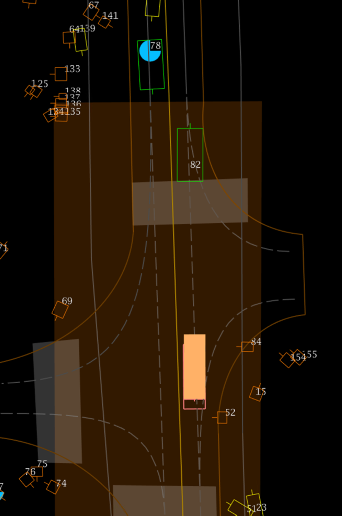}
		\includegraphics[height=\eh]{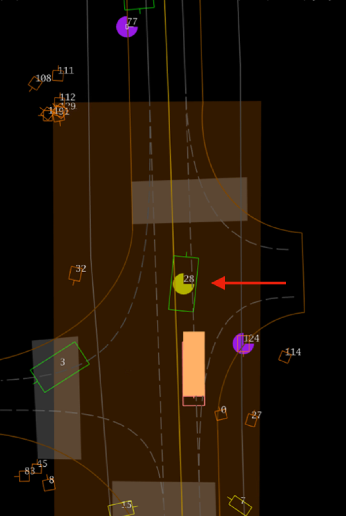}
		\includegraphics[height=\eh]{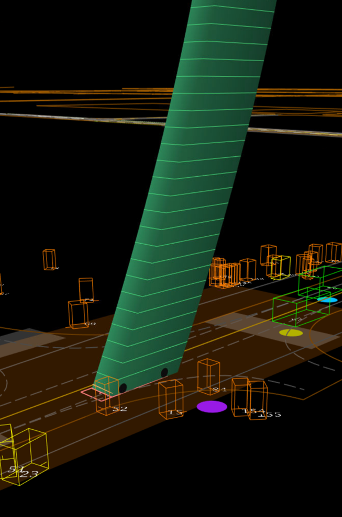}
		\includegraphics[height=\eh]{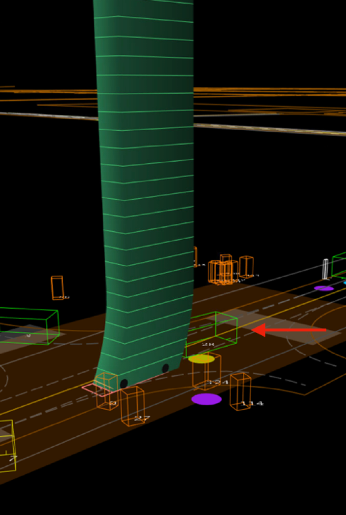}
		\caption{\figcapmaker{Illustration of AV behaviours in reaction to ground truth and actual
		noisy perception inputs~(best viewed in colour).}
		Under the ground truth perception input (the first and the third pictures), the AV is clear to
		move forward with soft braking to keep distance to another vehicle (`82') in front.
		Given the noisy perception input (the second and the fourth pictures), however, 
		the AV has to hard brake
		to avoid a potential collision with the false positive vehicle (`28') close to it in front (marked
		 by the red arrow). In both cases, since the AV speed is slow and is braking (either
		 soft or hard), the difference in behaviour is insignificant (PKL = -0.802), yet the
		consequence of the false positive is by no means trivial: the false positive causes a hard brake
		 and virtual collision (between the behaviours given ground truth perception input and false
		positive), which is precisely captured by the proposed TIP (-115.42). The kinematic motion
		for the ground truth scenario (bottom left) is $a = -0.36$m/s\textsuperscript{2}, $j = -0.72$m/s\textsuperscript{3}, and for
		the noisy scenario (bottom right) is $a = -0.36$m/s\textsuperscript{2}, $j = -76.4$m/s\textsuperscript{3}. {\bf Note how sharp
		the braking changes in presence of the noisy perception} (jerk: $-0.72$m/s\textsuperscript{3} versus $-76.4$m/s\textsuperscript{3}).
		Clearly, this is a critical error from the system's perspective.
		}
		\label{fig:prt_demo}
\end{figure*}

\begin{figure}[t]
\def \eh{2.4in}
		\centering
		\includegraphics[height=\eh]{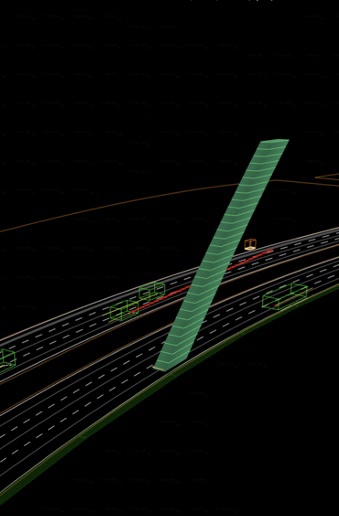}
		\includegraphics[height=\eh]{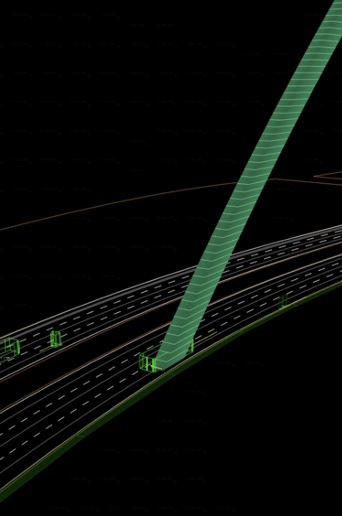}
		\includegraphics[height=\eh]{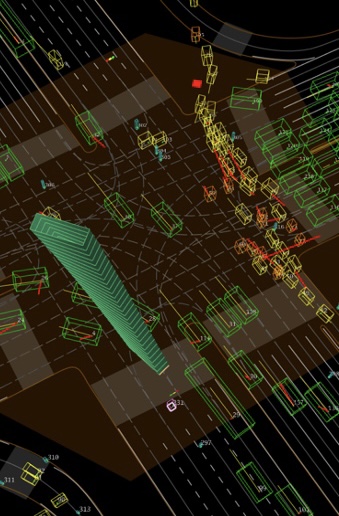}
		\includegraphics[height=\eh]{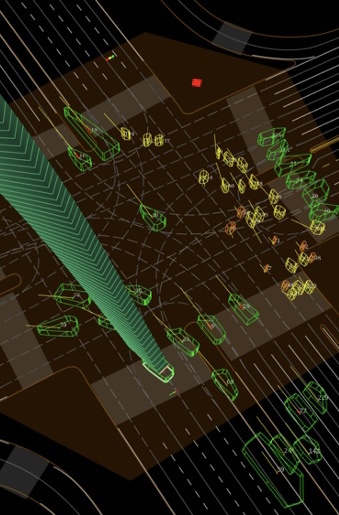}
		\caption{\figcapmaker{Illustration of more AV behaviours in reaction to ground truth and actual noisy
		perception inputs~(best viewed in colour).}
		Two more outrageous perception errors are shown where an object location is improperly
		 perceived such that it is superimposed with the AV.
		The ground truth is shown in the first and third pictures and the actual noisy perception in 
            the second and fourth, respectively.
		For the first case: TIP = -132.4, PKL = 0.0. 
		For the second case: TIP = -75.0, PKL = 0.04. 
		}
		\label{fig:prt_demo2}
\end{figure}

 {\bf Results on Real Data.}
On the real data, we also have similar observations, which is demonstrated by an actual scene for one such scenario in~\reffigure{fig:prt_demo}:
a falsely detected vehicle in front of the AV does not change the AV's behaviour considerably (PKL = -0.802), while the significant planning cost change is reflected by TIP with a value -115.42. More individual examples are shown in Figure~\ref{fig:prt_demo2}.

Overall, on both synthetic and real data, the proposed TIP is shown to efficiently and effectively capture perception errors critical to AV planning that may be missed by PKL.
This confirms our motivation to exploit the actual AV planning process, as opposed to the planning result only, to gain insights into the impact of input perception error on AV driving quality.

%%%%%%%%%%%%%%%%%%%%%%%%%%%%%%
%%%%%%%%%%%%%%%%%%%%%%%%%%%%%%

%%%%%%%%%%%%%%%%%%%%%%%%%%%%%%
%%%%%%%%%%%%%%%%%%%%%%%%%%%%%%

\subsection{Comparison with IPA}

Injecting planning-awareness (IPA) is recently proposed by~\citep{Ivanovic2021} to encode the planning error based on the hypothesis that the impact of an object location error is proportional to the gradient magnitude of the planning cost functions involving the AV-object distance. 
This solution requires differentiability of the planning cost functions, while our approach does not and thus is more applicable to a modularised SAE Level 4 AV that typically comprises a pipeline of individual components including perception~\citep{simtrack}, prediction~\citep{prophnet}, planning~\citep{Bronstein2022}, \etc~Even more serious, IPA fails to account for all cases since the local properties (gradients)
 do not always reflect the global ones (overall losses).
To illustrate this, consider a scenario, where the cost of AV being close to an object is $1/d$. 
Now assume that there are two cases of object location errors.
\begin{itemize}

\item Case one:
The ground truth distance of an object to the AV is 1m, and the noisy distance estimated by perception is 0.9m.
Per IPA defined in~\citep{Ivanovic2021}, the result is
$$
\abs{\left.\frac{\dif}{\dif d}(1/d)\right|_{d=1}} \abs{\Delta d} = 1 \times \abs{1.0-0.9} = 0.1,
$$
while the actual cost difference is $\abs{\frac{1}{0.9}-\frac{1}{1}}=1/0.9-1=0.111$.

\item Case two:
The ground truth distance of an object to the AV is 2m, and the noisy distance estimated by perception is 2.5m.
Per IPA defined in~\citep{Ivanovic2021}, the result is
$$
\abs{\left.\frac{\dif}{\dif d}(1/d)\right|_{d=2}} \abs{\Delta d} = 0.25 \times \abs{2.5-2.0} = 0.125,
$$
while the actual cost difference is $\abs{\frac{1}{2}-\frac{1}{2.5}}=0.5-1/2.5=0.1$.
\end{itemize}
Obviously, IPA score of case two is larger than case one, while the actual error in planning cost
 is the other way, as the Taylor series up to first-order terms adopted by IPA cannot precisely
 delineate the cost function value change over a large input variation.

\section{Application to Neural Planners}
\label{app:neuralplanner}

%%%%%%%%%%%%%%%%%%%%%%%%%%%%%%
\begin{figure*}[t]
		\centering
		\includegraphics[width=\linewidth]{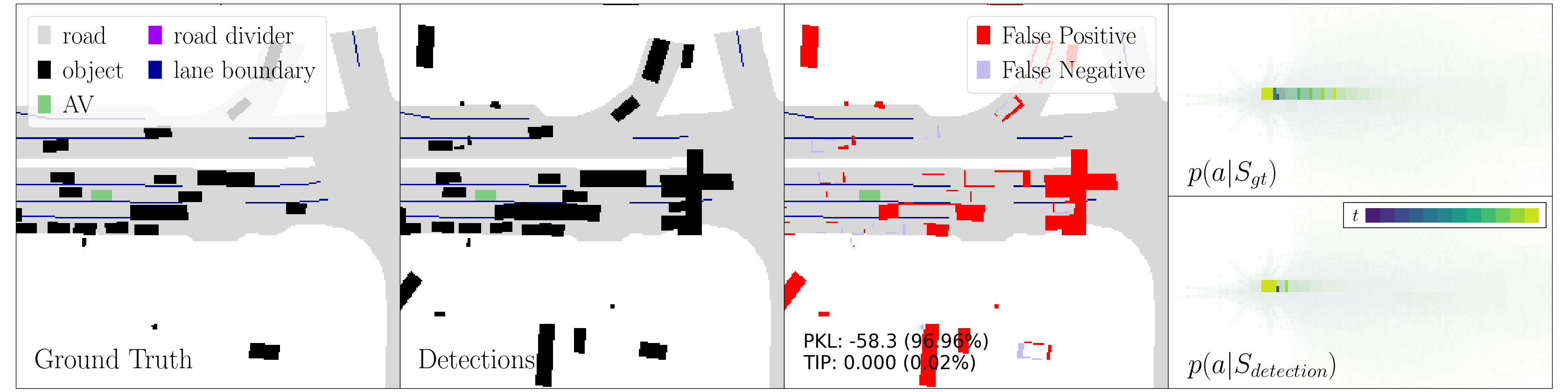}
		\includegraphics[width=\linewidth]{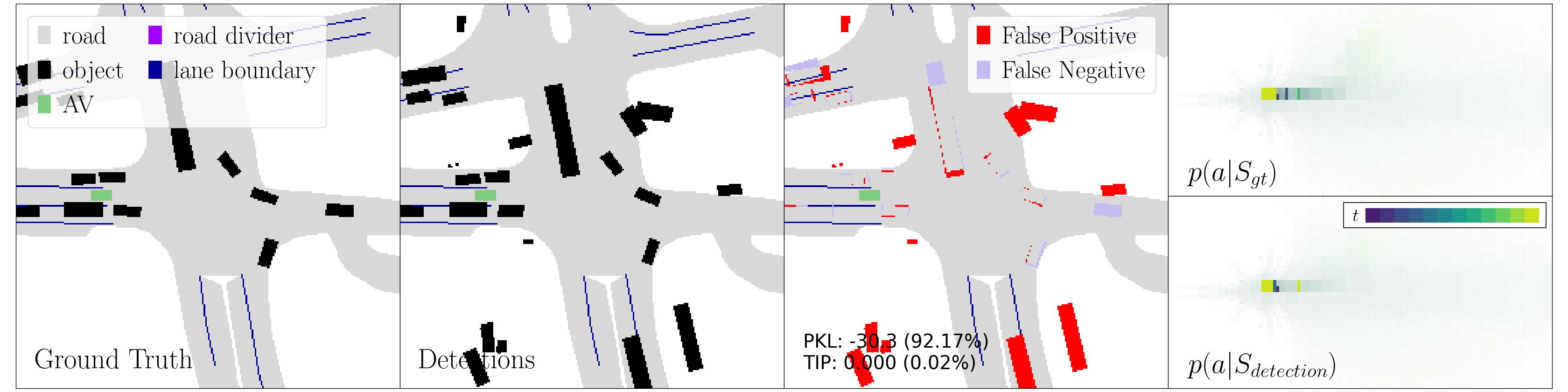}
		\includegraphics[width=\linewidth]{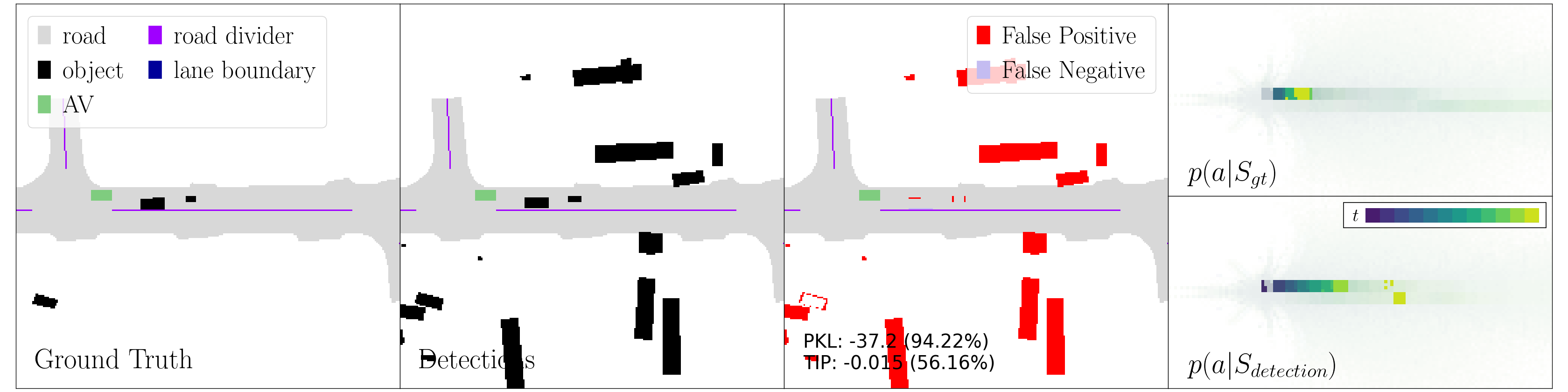}
		\caption{
		{\figcapmaker{Detection results and metric scores on nuScenes dataset (best viewed in colour).} From left to right columns: (i) ground truth annotations; (ii) detection results; (iii) difference between ground truth and detection results, with PKL and TIP scores for the scenario (score percentiles in the whole dataset are also shown in parentheses); and (iv) AV location distributions $\probb(\action|\staterv)=\probb(\{l_t\}|\staterv)$ at different time steps predicted by the planner, with one colour map corresponding to the location distribution at one time step (the action distribution density has been enhanced for visual saliency and not plotted to the numeric scale, and the most likely positions at each time step are marked with the corresponding solid colour).}
		}
		\label{fig:nuscenes}
\end{figure*}
%%%%%%%%%%%%%%%%%%%%%%%%%%%%%%

Following the discussion in \refsection{sec:neural}, more scenarios where PKL and TIP scores differ are illustrated in~\reffigure{fig:nuscenes} for detection results by the CBGS detector on nuScenes validation dataset. 
A typical observation is that, for the neural planner employed, when the 
optimal AV action (subject to kinetic and kinematic constraints) is to remain stationary 
regardless of the input 
perception noises (the first two examples in~\reffigure{fig:nuscenes}), 
TIP generally predicts an insignificant impact of the error while PKL may be 
dominated by the difference in low-probability regions where the KL-divergence is 
considerable (note that the result $p\log\frac{p}{q}$ could be large for any given 
$p>0$ when $q\rightarrow0$).
For similar reasons, PKL also tends to overestimate the impact in some cases where the AV is not stationary (the third example in~\reffigure{fig:nuscenes}).    

\begin{figure*}[t]
		\centering
		\includegraphics[width=0.24\linewidth]{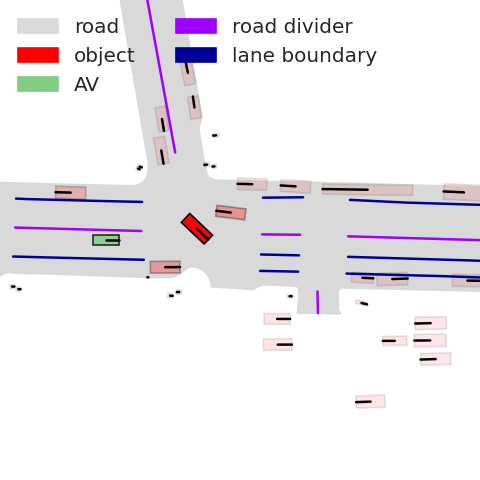}
		\includegraphics[width=0.24\linewidth]{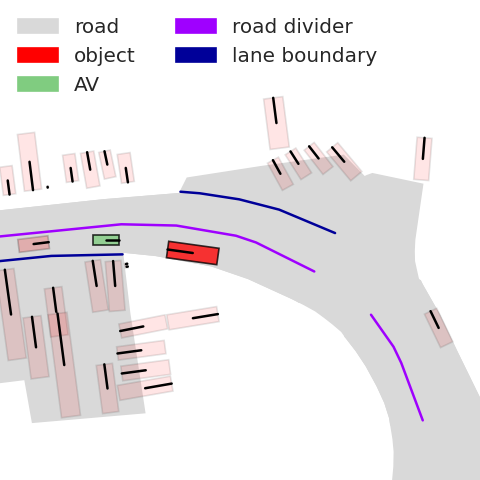}
		\includegraphics[width=0.24\linewidth]{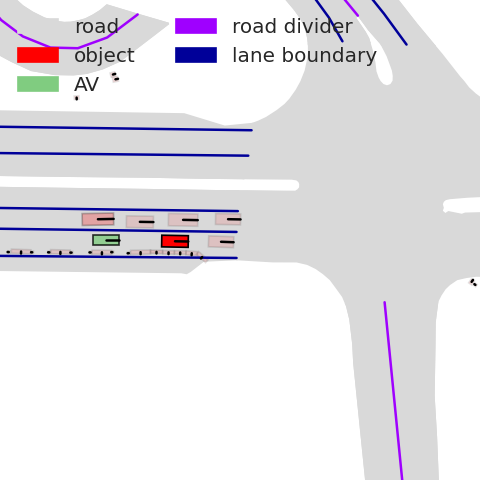}
		\includegraphics[width=0.24\linewidth]{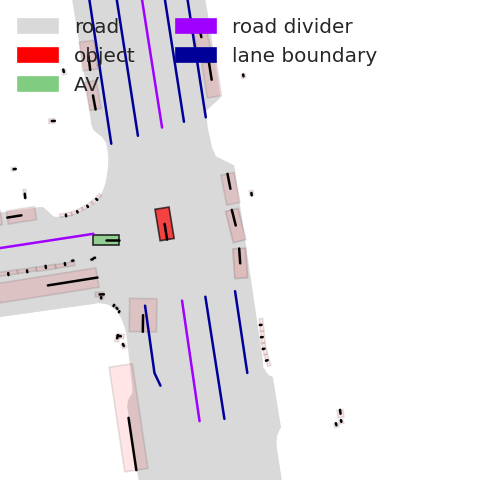}
		\includegraphics[width=0.24\linewidth]{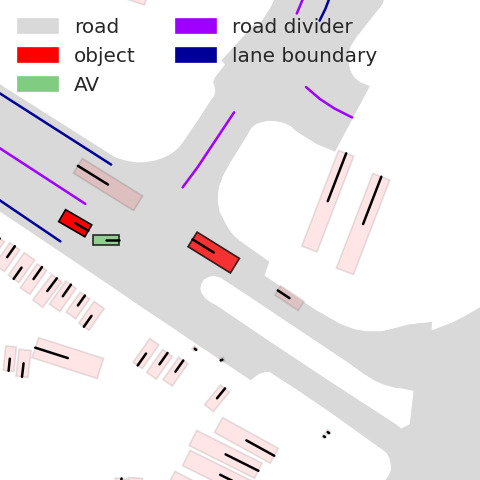}
		\includegraphics[width=0.24\linewidth]{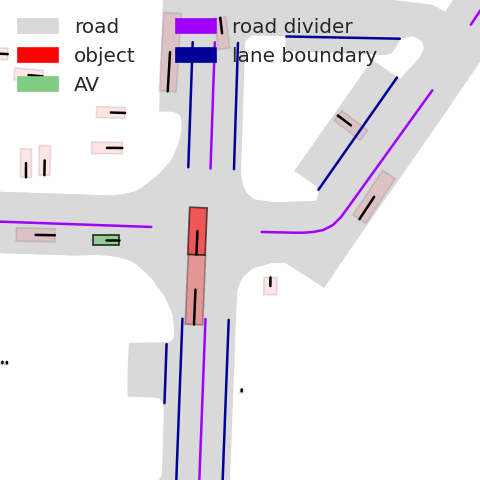}
		\includegraphics[width=0.24\linewidth]{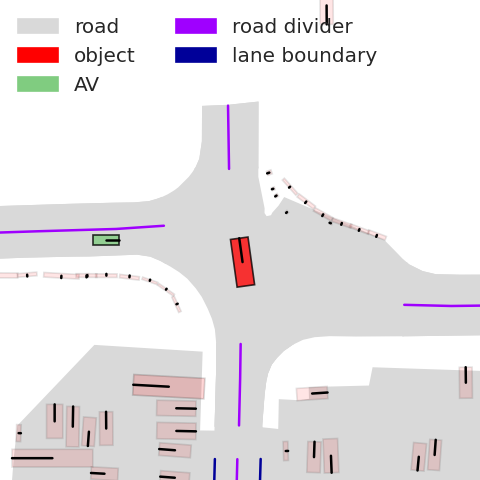}
		\includegraphics[width=0.24\linewidth]{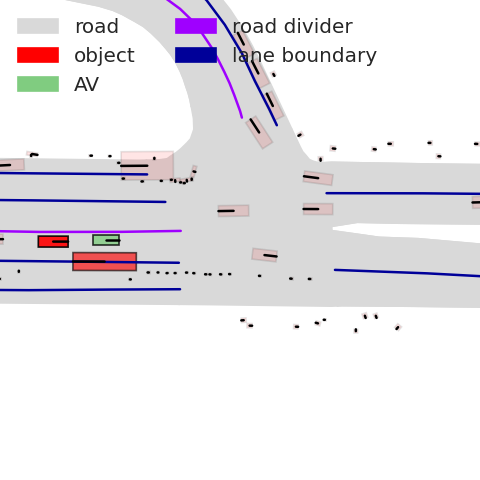}
		\caption{
		{\figcapmaker{Examples of crucial false negatives predicted by TIP (best viewed in colour).} The colour saturation indicates the significance of the error if the corresponding vehicle is missed by the detector.
		}
		}
		\label{fig:fneg_demo}
\end{figure*}

\begin{figure*}[h]
		\centering
		\includegraphics[width=0.24\linewidth]{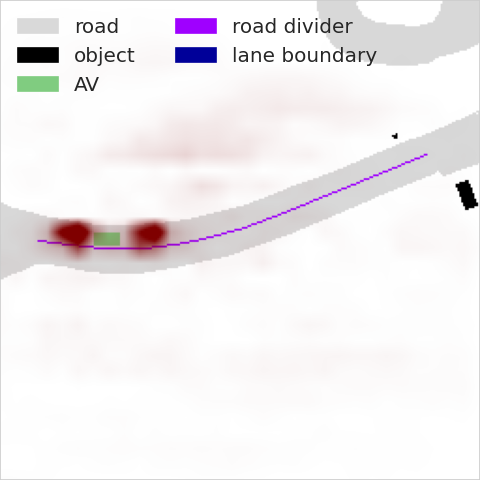}
		\includegraphics[width=0.24\linewidth]{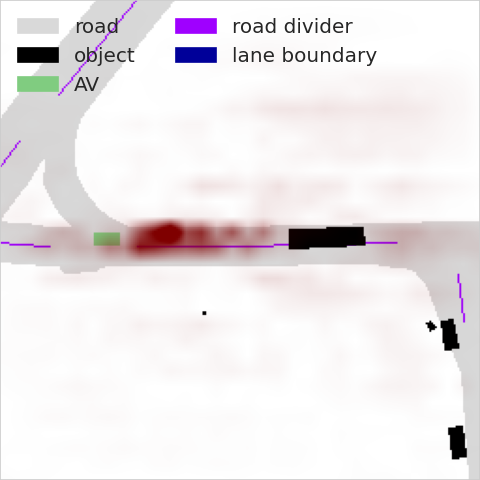}
		\includegraphics[width=0.24\linewidth]{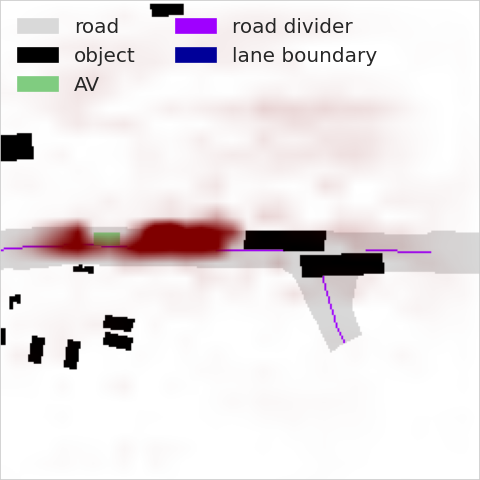}
		\includegraphics[width=0.24\linewidth]{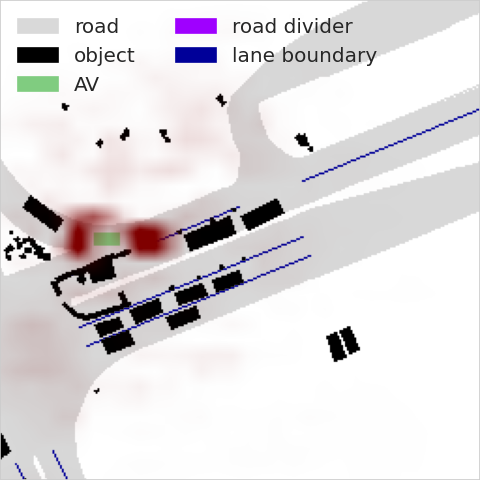}
		\includegraphics[width=0.24\linewidth]{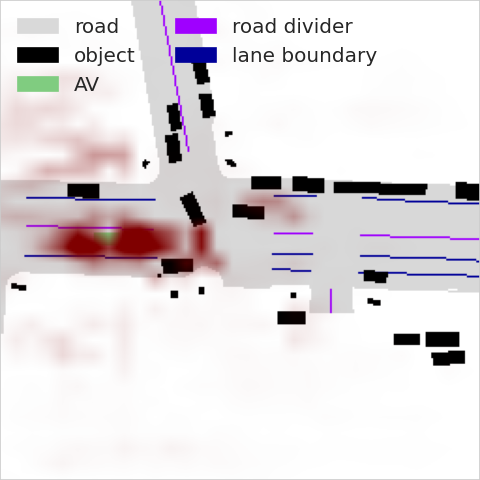}
		\includegraphics[width=0.24\linewidth]{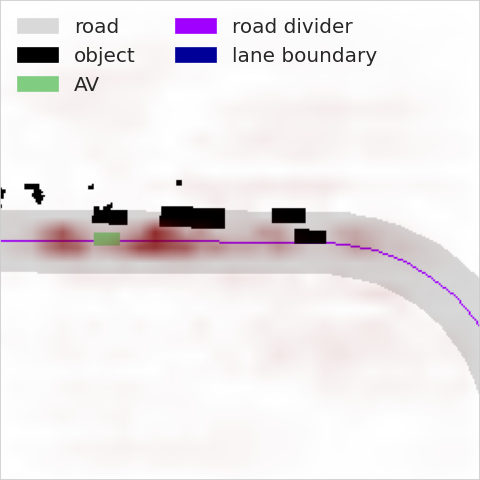}
		\includegraphics[width=0.24\linewidth]{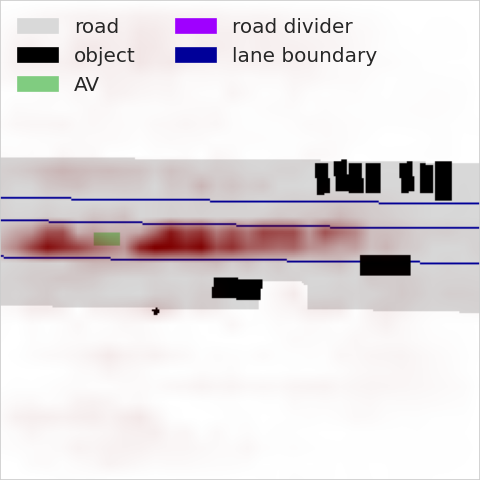}
		\includegraphics[width=0.24\linewidth]{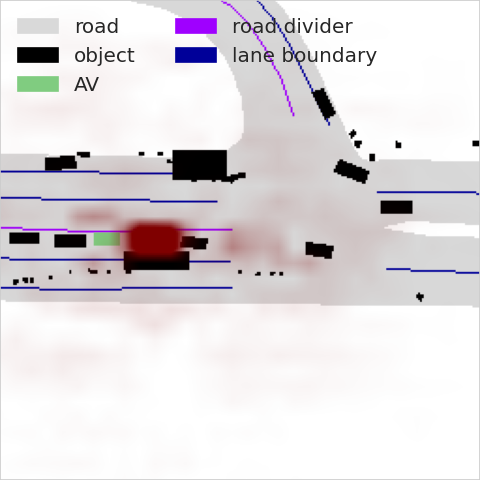}
		\caption{
		{\figcapmaker{Examples of crucial false positives predicted by TIP (best viewed in colour).} The colour saturation indicates the significance of the error if a ghost vehicle is falsely detected at the corresponding location.
		}
		}
		\label{fig:fpos_demo}
\end{figure*}

In addition to scoring a particular detection result, the proposed metric can also predict
 sensitive regions where false positives or true positives are most crucial for a planner.
For this, we measure the impact of false negatives by removing vehicles from the ground truth 
annotations and evaluating the TIP score of the synthetic scene, with results 
presented in~\reffigure{fig:fneg_demo}.    
Similarly, the significance of false positives is predicted by adding a ghost vehicle
 at a location and evaluating the TIP score of the scene, which is illustrated in~\reffigure{fig:fpos_demo}.  
Overall, the most critical false positives or negatives are identified at the locations along the future spatiotemporal 
path of the AV that require AV-object interaction.
Interestingly, the neural planner may reverse in some cases, producing nontrivial TIP scores 
for false positives or negatives behind it.
This observation is distinct from that in~\refsection{sec.exp.sync.case}, where a false negative 
behind the AV has no impact on the AV planner that does not reverse.      
  
%%%%%%%%%%%%%%%%%%%%%%%%%%%%%%
%%%%%%%%%%%%%%%%%%%%%%%%%%%%%%
%%%%%%%%%%%%%%%%%%%%%%%%%%%%%%
%%%%%%%%%%%%%%%%%%%%%%%%%%%%%%

\section{Examples and Non-Examples of Square-Integrable Density Functions}
\label{app:example}

\refthm{thm.hilbert} in the main text requires square-integrability of a density function,
which includes many popular cases that may be used for constructing the utility function for planning.
\begin{example}[Bounded \pdf s]
If both the support and range of the \pdf~$\pdfs(x)$ of a random variable are bounded,
    then $\pdfs(x)$ is square-integrable, \eg~uniform distribution.
\qed
\end{example}

\begin{example}[Parametric \pdf s]
\pdf s of many popular parametric statistical models are square-integrable, \eg~
(sub-)Gaussian, (sub-)Laplace, Gamma (including exponential, Erlang, and
$\chi^2$ distribution), \etc
\qed
\end{example}

\begin{example}[Mixture Models of Countable Components with Square-Integrable \pdf s]
The \pdf~of a mixture model is of the form:
    \begin{equation}
    \label{eq:mix.pdf}
    \pdfs(x) = \sum\nolimits_{i} \alpha_i\pdfs_i(x),~\alpha_i>0,~\sum\nolimits_{i}\alpha_i=1,
    \end{equation}
    where $\pdfs_i(x)$ is the \pdf~of the $i$-th component out of the {\it countable} set $\{f_i(x)\}$.
    $\pdfs(x)$ of \refeqn{eq:mix.pdf} is square-integrable if $\forall i, \pdfs_i\in L^2$ and $M=\sup_i \norm{\pdfs_i}_\Hspace<+\infty$ as
    \begin{align}
    \int\abs{\pdfs(x)}^2\dif x
    &= \int \sum_{i,j} \alpha_i\alpha_j\pdfs_i(x)\pdfs_j(x)\dif x
    = \sum_{i,j} \alpha_i\alpha_j \innerprod{\pdfs_i}{\pdfs_j}_\Hspace\\
    &\leqslant \sum_{i,j} \alpha_i\alpha_j  \norm{\pdfs_i}_\Hspace\norm{\pdfs_j}_\Hspace
    \leqslant M^2
    <+\infty. \nonumber
    \end{align}
    A variety of mixture models are included
    such as Gaussian mixture models and mixtures of Gamma distribution.
\qed
\end{example}

On the other hand, since $\Lnorm{1}$ and $\Lnorm{2}$ norms are not necessarily equivalent in
infinite-dimensional spaces, 
there are indeed some density functions $\pdfs(x)\in L^1$ with
infinite $\Lnorm{2}$ norm.
\begin{nonexample}[Square-Unintegrable \pdf s]
Let the distribution $\distrib_X$ of a random variable $X$ be
$$
\distrib_X(x) =
    \begin{cases}
    0, &x\in(-\infty, 0)\\
    \frac{1}{\sqrt{a}}x^{\frac{1}{2}}, &x\in[0, a]\\
    1, &x\in(a, +\infty)
    \end{cases}
$$
where $a>0$ is the parameter;
and the density function is then
    $$
    \pdfs(x) =
    \begin{cases}
    \frac{1}{2\sqrt{a}}x^{-\frac{1}{2}}, &x\in(0, a)\\
    0, &\textnormal{otherwise}
    \end{cases}
    $$
where $\pdfs(x)$ is not square-integrable since $x^{-1}$ increases too fast as $x\rightarrow0$.
\qed
\end{nonexample}

\section{Proofs of Theorems in the Main Text}
\label{app:proof}

\subsection{Notations}
Besides the notations in~\refsection{sec:preliminary}, a few more are introduced as follows.
A unit step function is $\funcStep(x-c)=\1{x\in[c, +\infty)}, c\in\real$.
$L^1(\sspace, \measure)$ denotes the space of absolutely integrable functions.

\subsection{Embedding Probability Measures in $\Hspace$}

\begin{myproof}[\refthm{thm.hilbert}]
Since $\distrib_X(x)$ is absolutely continuous, there exists a density function $\pdfs_X(x)\in L^1$
such that
\begin{equation}
   \frac{\dif}{\dif x}\distrib_X(x) = \pdfs_X(x)
\end{equation}
almost everywhere.
Since $\pdfs_X(x)\in L^2$ , let
$M=\norm{\pdfs_X}<+\infty$,
$\forall g\in\Hspace$,
we have
\begin{align}
   \abs{\EVsup{X}{g(x)}}
   =&~ \abs{\int_xg(x)\dif F_X(x)} \\
   =&~ \abs{\int_xg(x)\pdfs(x)\measure(\dif x)} \\
   \leqslant&~ \int_x\abs{g(x)}\abs{\pdfs(x)}\measure(\dif x)  \\
   \leqslant&~ M\normp{g}{\Hspace}, \label{eq:Ev.bound2}
\end{align}
where \refeqn{eq:Ev.bound2} follows from the Cauchy-Schwarz inequality~\citep[Theorem 11.35]{Rudin1976pma}.
Thus, the linear functional $\EVsup{X}{\cdot}$ is bounded on $\Hspace$ and
$$
\EVsup{X}{g(x)}
=\int_xg(x)\dif \distrib_X(x)
=\int_x\pdfs_X(x)g(x)\measure(\dif x)
=\innerprod{\pdfs_X}{g}_\Hspace,~\forall g\in\Hspace,
$$
where $\mu_\probb\coloneqq\pdfs_X\in\Hspace$ is the embedding of the probability measure
in $\Hspace$.
Now assume that there exists another element $\mu'\in\Hspace$ such that
$$
\EVsup{X}{g(x)}=\innerprod{\mu'}{g}_\Hspace,~\forall g\in\Hspace.
$$
Since $\mu_\probb-\mu'\in\Hspace$, we have
\begin{align}
\norm{\mu_\probb-\mu'}^2_\Hspace
=&~ \innerprod{\mu_\probb-\mu'}{\mu_\probb-\mu'}_\Hspace \nonumber\\
=&~ \innerprod{\mu_\probb}{\mu_\probb-\mu'}_\Hspace - \innerprod{\mu'}{\mu_\probb-\mu'}_\Hspace \nonumber\\
=&~ \EVsup{X}{\mu_\probb-\mu'} - \EVsup{X}{\mu_\probb-\mu'} \nonumber\\
=&~ 0. \nonumber
\end{align}
Therefore, the embedding $\mu_\probb$ for probability measure
$\probb$ in $\Hspace$ is a unique equivalence class of the functions that are equal almost everywhere.
\qed
\end{myproof}

\subsection{Injection of Probability Measure Embeddings in $\Hspace$}
To prove the injection of probability measure embedding in
\refthm{thm.hilbert.injection},
a preliminary result is first introduced.
\begin{lemma}[Lemma 9.3.2 of \citep{dudley2002}]
\label{lemma:prob.equality}
If $(\calspace{X}, \metricD)$ is a metric space, $\probb$ and $\qrobb$ are two probability measures on $\calspace{X}$, then $\EVsup{x\sim\probb(x)}{g} = \EVsup{x\sim\qrobb(x)}{g}, \forall g\in C_b(\calspace{X})$ if and only if $\probb=\qrobb$, where $C_b(\calspace{X})$ is the space of all bounded continuous functions on $\calspace{X}$.
\end{lemma}
\begin{myproof}[\refthm{thm.hilbert.injection}]
Now we prove this theorem in the following two directions.

{\bf Necessity}.
Since the embedding of a probability measure is unique in $\Hspace$, it is easy to see that $\mu_\probb=\mu_\qrobb$ if $\probb=\qrobb$.

{\bf Sufficiency}.
Note that, by Weierstrass extreme value theorem~\citep[Theorem 4.16]{Rudin1976pma}, any real continuous function $g\in C(\sspace)$ on the compact space $\sspace$ is bounded, \ie
$\forall g\in C(\sspace), \exists M\in\real$ such that $\abs{g(x)}<M, \forall x\in\sspace$.
It follows that $C(\sspace)\subset L^2(\sspace)$ since
$$
\int_\sspace\abs{g(x)}^2\measure(\dif x) \leqslant M^2\abs{\sspace} < +\infty.
$$
Now if $\mu_\probb=\mu_\qrobb$ almost everywhere, we have
\begin{align}
\abs{\EVsup{\probb}{g(x)} - \EVsup{\qrobb}{g(x)}}
&= \abs{\innerprod{\mu_\probb}{g} -   \innerprod{\mu_\qrobb}{g}}
= \abs{\innerprod{\mu_\probb-\mu_\qrobb}{g}} \\
&\leqslant \norm{\mu_\probb-\mu_\qrobb}_\Hspace  \norm{g}_\Hspace
= 0,~\forall g\in C(\sspace).
\end{align}
Thus $\probb=\qrobb$ by~\reflemma{lemma:prob.equality}.
\qed
\end{myproof}

\subsection{Approximation of Expectation for Discrete/Mixed distribution in $\Hspace$}
While Theorem 1 in the main text only addresses the continuous distribution,
a similar result can be found given
point-wise continuity conditions for general distribution, which
can be decomposed into absolutely continuous and discrete parts~\citep{chung2000course}.
\setcounter{theorem}{3}
\begin{theorem}[Approximation of Mixed Distribution]
\label{thm.hilbert.mixed}
Let $\distrib_{ac}(x)$ be an absolutely continuous distribution function with density function $\pdfs_X(x)$;
$\distrib_d(x)=\sum_ib_i\funcStep(x-a_i)$ a discrete distribution function of point mass
at a countable set $\{a_i\}$ such that $b_i>0$ and $\sum_ib_i=1$;
$\distrib_X(x)=\lambda\distrib_{ac}(x)+(1-\lambda)\distrib_d(x)$ a mixed distribution function with $\lambda\in(0,1)$ as the convex combination coefficient.
If $\pdfs_X(x)$ is square-integrable,
and $g(x)\in L^2$ is uniformly continuous at $\{a_i\}$,
then there exists a sequence of $\{\mu_{\probb, n}\}\subset\Hspace$ such that
\begin{equation}
\lim_{n\rightarrow\infty}\innerprod{{\mu_{\probb, n}}}{g}_{\Hspace}
=
\EVsup{X}{g(x)}.
\end{equation}
\end{theorem}

We start by considering a simple discrete case by the following lemma.
\begin{lemma}
\label{thm.hilbert.discrete}
Let $\distrib_X(x)=\funcStep(x-a)$ be a discrete distribution function with point mass at $a\in\sspace$.
If $g(x)\in L^2$ is continuous at $a$, then there exists a sequence of $\{\mu_{\probb, n}\}\subset\Hspace$ such that
\begin{equation}
\lim_{n\rightarrow\infty}\innerprod{{\mu_{\probb, n}}}{g}_{\Hspace}
=
\EVsup{X}{g(x)}.
\end{equation}
\end{lemma}
\begin{myproof}[\reflemma{thm.hilbert.discrete}]
$\forall\varepsilon>0$,
since $g(x)$ is continuous at $a$, there exists a
a radius $r>0$ such that
$$
g(a)-\varepsilon \leqslant g(x) \leqslant g(a)+\varepsilon, ~\forall x\in B(a,r)
$$
with a positive measure $V=\measure(B(a,r))>0$,
where $B(a,r)\subset\sspace$ is a neighbourhood of $r$ around $a$.
Define
$$
h_\varepsilon(x) = \frac{1}{V}\1{x\in B(a,r)} \in \Hspace.
$$
We have
$$
g(a)-\varepsilon < \innerprod{h_\varepsilon}{g}_{\Hspace} < g(a)+\varepsilon.
$$
Thus,
$$
\lim_{n\rightarrow\infty}\innerprod{{h_{\frac{1}{n}}}}{g}_{\Hspace}
= g(a)
= \EVsup{X}{g(x)}.
\footnote{$\{h_{\frac{1}{n}}\}$ itself, however, is not a Cauchy sequence, thus it has no limit.}
$$
\qed
\end{myproof}
\reflemma{thm.hilbert.discrete} implies that the expected value of a function continuous
at the point mass of a delta distribution can be approximated by an inner
product in $\Hspace$ with any {\it arbitrary precision}.

\begin{myproof}[\refthm{thm.hilbert.mixed}]
Note that
$$
\EVsup{X}{g(x)}=\lambda\int_x g(x)\dif \distrib_{ac}(x) + (1-\lambda)\sum\nolimits_i b_ig(a_i).
$$
Since $\distrib_{ac}(x)$ is absolutely continuous, by
Theorem 1, there exists a $\mu\in\Hspace$ such that
\begin{equation}
\int_x g(x)\dif \distrib_{ac}(x) = \innerprod{\mu}{g}_\Hspace,~\forall h\in\Hspace.
\label{eq:app.proof.mixed.cont}
\end{equation}
On the other hand, $\forall\varepsilon>0$, since $g(x)$ is {\it uniformly continuous} at $\{a_i\}$,
 there exists a radius $r>0$ such that
 $\forall i$,
$$
g(a_i)-\varepsilon < g(x) < g(a_i)+\varepsilon, ~\forall x\in B(a_i,r)
$$
and $V=\measure(B(a_i, r))>0$ (translation invariance of Lebesgue measures in $\real^d$).
Define
$$
h_\varepsilon(x) = \frac{1}{V}\sum\nolimits_ib_i\1{x\in B(a_i,r)} \in \Hspace.
$$
We have
$$
\sum\nolimits_i b_i g(a_i)-\varepsilon
= \sum\nolimits_i b_i g(a_i)-\varepsilon\sum\nolimits_i b_i
< \innerprod{h_\varepsilon}{g}_{\Hspace}
< g(a)+\varepsilon\sum\nolimits_i b_i
= \sum\nolimits_i b_i g(a_i)+\varepsilon,
$$
Thus
\begin{equation}
\lim_{n\rightarrow\infty}\innerprod{{h_\frac{1}{n}}}{g}_{\Hspace}
= \sum\nolimits_i b_i g(a_i).
\label{eq:app.proof.mixed.discrete}
\end{equation}
Combining \refeqn{eq:app.proof.mixed.cont} and \refeqn{eq:app.proof.mixed.discrete} leads to
$$
\lim_{n\rightarrow\infty}\innerprod{{\lambda\mu+(1-\lambda)h_\frac{1}{n}}}{g}_{\Hspace}
= \lambda\int_x g(x)\dif \distrib_{ac}(x) + (1-\lambda)\sum\nolimits_i b_ig(a_i)
= \EVsup{X}{g(x)}.
$$
\qed
\end{myproof}

\subsection{Uniform Convergence Rate of Expected Utility Estimators}

\begin{myproof}[\refthm{thm.estimator.bound.exp}]
Assume that $\{X_i\}_{i=1}^{n}$ and independent and $X_i\in[a_i, b_i]$ almost surely.
Let $\bar{X}=\frac{1}{n}\sum_iX_i$.

Per Hoeffding's inequality \citep[Theorem 2]{Hoeffding1963},
for any $\varepsilon>0$,
\begin{equation}
\label{eq:estimator.eu.hoeffding1}
\Prob\Big(\bar{X}-\EVs{\bar{X}}>\varepsilon\Big)
< \exp{\bigg\{-\frac{2n^2\varepsilon^2}{\sum_{i=1}^n(b_i-a_i)^2}\bigg\}}.
\end{equation}
By symmetry, it also holds true that, for any $\varepsilon>0$,
\begin{equation}
\label{eq:estimator.eu.hoeffding2}
\Prob\Big(\bar{X}-\EVs{\bar{X}}<-\varepsilon\Big)
< \exp{\bigg\{-\frac{2n^2\varepsilon^2}{\sum_{i=1}^n(b_i-a_i)^2}\bigg\}}.
\end{equation}
Combining one-side inequalities of \refeqn{eq:estimator.eu.hoeffding1} and \refeqn{eq:estimator.eu.hoeffding2}
 leads to
\begin{equation}
\label{eq:estimator.eu.hoeffding}
\Prob\Big(\abs{\bar{X}-\EVs{\bar{X}}}>\varepsilon\Big)
< 2\exp{\bigg\{-\frac{2n^2\varepsilon^2}{\sum_{i=1}^n(b_i-a_i)^2}\bigg\}}
\leqslant  2\exp{\bigg(-\frac{n\varepsilon^2}{2M^2}\bigg)},~\forall\varepsilon>0,
\end{equation}
where $M=\sup(\{\abs{a_1}, \cdots, \abs{a_n}, \abs{b_1}, \cdots, \abs{b_n}\})$.

On the other hand, Bernstein inequality~\citep{Bernstein1946} also provides an improved
revision of Chebyshev's inequality by incorporating both almost-sure bound and variance bound:
\begin{equation}
\label{eq:estimator.eu.bernstein}
\Prob\bigg(\abs{\bar{X}-\EVs{\bar{X}}}>\varepsilon\bigg)
< 2\exp\bigg\{-\frac{n\varepsilon^2}{2\var(\bar{X})+2M\varepsilon/3}\bigg\},
~\forall\varepsilon>0.
\end{equation}
The proof is completed by setting $X_i=\Utility(\staterv_i, \action)$ and
taking the lowest bound of \refeqn{eq:estimator.eu.hoeffding} and \refeqn{eq:estimator.eu.bernstein}
for the tail probability of $\abs{\bar{X}-\EVs{\bar{X}}}$.
\qed
\end{myproof}

\end{document}